\documentclass[final,3p,times]{elsarticle}

\usepackage{amsmath,amssymb,amsfonts}
\usepackage{mathtools}
\newtheorem{definition}{Definition}
\newtheorem{example}{Example}

\newtheorem{remark}{Remark}
\newtheorem{theorem}{Theorem}
\newtheorem{proof}{Proof}

\usepackage{graphicx}
\usepackage{array}
\usepackage{multirow}
\usepackage{makecell}
\usepackage[caption=false,font=footnotesize,labelfont=sf,textfont=sf]{subfig}

\usepackage{braket}
\usepackage{qcircuit}
\usepackage{stfloats}  
\usepackage{float}
\usepackage{url}       
\usepackage{adjustbox}
\usepackage{indentfirst}
\usepackage{tikz}
\usetikzlibrary{shapes.geometric,arrows,positioning,calc,fit,decorations.pathreplacing}

\usepackage[linesnumbered,ruled,vlined]{algorithm2e}

\usepackage[normalem]{ulem}
\usepackage{xcolor}
\definecolor{qWire}{RGB}{66,66,66}
\definecolor{qCtrl}{RGB}{42,42,42}
\definecolor{qXDraw}{RGB}{43,126,64}
\definecolor{qXFill}{RGB}{232,248,237}
\definecolor{qRyDraw}{RGB}{55,94,185}
\definecolor{qRyFill}{RGB}{235,240,255}
\definecolor{qOpDraw}{RGB}{133,79,173}
\definecolor{qOpFill}{RGB}{248,241,252}
\definecolor{qMeasDraw}{RGB}{0,125,138}
\definecolor{qMeasFill}{RGB}{229,247,249}

\allowdisplaybreaks[1]

\usepackage[utf8]{inputenc}

\usepackage[hidelinks]{hyperref}

\newcommand{\circledsmall}[1]{\hbox{\tikz\draw (0pt, 0pt)
		circle (.5em) node {\makebox[0.15em][c]{\scriptsize#1}};}}
\newcommand{\circledtiny}[1]{\hbox{\tikz\draw (0pt, 0pt)
		circle (.4em) node {\makebox[0.01em][c]{\tiny#1}};}}
\newcommand{\circledlarge}[1]{\hbox{\tikz\draw (0pt, 0pt)
		circle (.6em) node {\makebox[0.15em][c]{\small#1}};}}

\journal{Information Fusion}

\begin{document}

\begin{frontmatter}



\title{Quantum Information Fusion and Correction under the Transferable Belief Model}

\author[label1]{Qianli Zhou\corref{cor1}}
\ead{zhou\_qianli@nwpu.edu.cn}
\cortext[cor1]{Corresponding author.}
\affiliation[label1]{organization={School of Electronics and Information, Northwestern Polytechnical University},
             city={Xi'an},
             postcode={710072},
             country={China}}
\author[label2]{Hao Luo}
\author[label3]{Lipeng Pan}
\author[label4]{Yong Deng}
\author[label5,label6]{{\' E}loi Boss{\' e}} 
	\affiliation[label2]{organization={College of Future Information Technology, Fudan University},
	city={Shanghai},
	postcode={200438},
	country={China}}
	\affiliation[label3]{organization={College of Information Engineering, Northwest A\&F University},
		city={Yangling},
		postcode={712100},
		country={China}}
\affiliation[label4]{organization={Institute of Fundamental and Frontier Science, University of Electronic Science and Technology of China},
	city={Chengdu},
	postcode={611731},
	country={China}}

\affiliation[label5]{organization={Department of Image and Information Processing, IMT-Atlantique},
	city={Brest},
	postcode={29238},
	country={France}}

\affiliation[label6]{organization={Expertises Parafuse},
	city={Quebec},
	postcode={G1W 4N1},
	country={Canada}}

\begin{abstract}
	The transferable belief model (TBM), developed within Dempster-Shafer theory, represents ambiguity and partial ignorance through set-valued belief masses. Its classical operations, however, can grow combinatorially over the power set. We formulate TBM reasoning on quantum circuits using the mass function quantum state (MFQS). The resulting framework covers belief representation, credal-level fusion and correction, product-space operations, and the generalized Bayesian theorem. For prepared MFQS inputs, CCR and the entire $\alpha$-junction use element-wise gate modules whose logical counts are linear in the number of frame elements. These circuits avoid explicit operation-stage updates over all $2^n$ focal-set coordinates. A noisy multi-domain classifier-fusion example further demonstrates interpretable source fusion and improved robustness to gate-control noise. The results establish Dempster-Shafer structures as a semantic layer for quantum information processing beyond singleton-probability representations.
\end{abstract}


\end{frontmatter}



\section{Introduction}
Reasoning under uncertainty is essential to trustworthy and interpretable artificial intelligence. Dempster-Shafer (DS) theory, also known as belief function theory, represents evidence that supports sets of possible values. The transferable belief model (TBM) \cite{smets1994transferable} gives these belief functions a two-level semantics. At the credal level, evidence updates and transfers belief masses. At the pignistic level, the current belief state is converted into a decision probability. This structure accommodates unreliable testimony, ambiguity, and partial ignorance that singleton probabilities cannot express directly. Belief functions have been applied to multi-source information fusion \cite{yang2013evidential}, expert decision making \cite{zhao2024mase,li2026novel}, fault diagnosis \cite{liu2024evidential,xu2025interactive}, multimodel fusion \cite{zhang2024mixed,han2025difference}, and computer vision \cite{huang2024intergration,geng2024causal,huang2025deep}. Their main computational difficulty arises from assigning masses on the power set \cite{deng2024plausibility,barhoumi2025empirical}; the cost of classical belief operations can therefore grow rapidly with the frame size.

Quantum computing stores information in quantum states and extracts probabilities by measurement \cite{singh2025qxai}. This model has motivated quantum machine learning methods for high-dimensional representation and optimization \cite{shi2026qsyncfold}. Quantization alone does not guarantee algorithmic acceleration \cite{dong2023machine}, particularly on noisy intermediate-scale quantum devices \cite{chen2023complexity}. The information representation must also match the structure of quantum circuits \cite{shi2024qsan}. A probability model usually derives a proposition $A\subseteq\Omega$ by aggregating singleton events. A basic probability assignment (BPA)\footnote{Also referred to as basic belief assignment in TBM.} directly assigns normalized weights on $2^\Omega$. Associating each element of $\Omega$ with one qubit maps every focal set to a computational basis state and every belief mass to a measurement probability. This element-qubit correspondence, first emphasized in \cite{zhou2023bf}, underlies the representation used here.

Previous studies have connected quantum formalisms with Dempster-Shafer theory through quantum-like evidence theory \cite{xiao2023generalized,xiao2026adaptive}, quantum models of mass functions \cite{deng2023novel}, and interference prediction \cite{pan2023evidential}. Under element-qubit encoding \cite{zhou2023bf}, qubits represent frame elements, computational basis states represent focal sets, and measurement probabilities recover belief masses. The marginal probability of an element qubit then equals its contour value. This correspondence gives quantum information processing a belief-theoretic interpretation. It has supported quantum-state encodings of BPAs and belief operations based on quantum algorithms \cite{zhou2023bf} and variational quantum linear solvers \cite{luo2024variational}. Boolean-algebra combination rules and attribute-fusion evidential classifiers have also been implemented on quantum circuits \cite{luo2026attribute,chen2026trustworthyqenn}.

This paper develops a circuit framework for TBM reasoning, including belief representation, credal fusion and correction, and product-space operations. Element-qubit encoding preserves focal sets as directly addressable circuit objects. Controlled gates can therefore manipulate ambiguity, partial ignorance, and set-valued propositions at the focal-set level, before any reduction to singleton probabilities.

The same correspondence provides a belief function interpretation of quantum kernel methods. Quantum kernels encode data as high-dimensional feature states and evaluate similarities through state overlaps or kernel estimators \cite{havlicek2019supervised}. Many supervised quantum learning models can accordingly be viewed as linear models or kernel methods in quantum feature spaces, although their useful structure may extend beyond an implicit kernel \cite{jerbi2023quantum}. Because the computational basis $\{0,1\}^n$ is isomorphic to $2^\Omega$, the measurement distribution of a feature state defines a mass function on focal sets. Its quantum kernel measures similarity between the induced belief functions. Existing kernel models rarely state this Dempster-Shafer interpretation explicitly. The proposed framework exposes this evidential structure and supports belief queries, fusion, correction, and product-space operations with controlled gates. For prepared states and query-level outputs, operations on $n$ element qubits can update the induced evidence distribution without materializing a $2^n$-dimensional vector.

Section \ref{pre} reviews TBM and quantum computing. Section \ref{method1} defines the MFQS representation and its circuit implementation. Sections \ref{method2} and \ref{method3} develop credal-level and product-space operations, respectively. Section \ref{sec:noise_experiment} evaluates multi-domain fusion under gate-control noise. Section \ref{con} concludes the paper.

\section{Preliminaries}
\label{pre}
\subsection{Transferable belief model}

TBM interprets Dempster-Shafer theory as an agent's belief state at a given time \cite{smets1994transferable}. Its original formulation distinguishes belief functions from probabilities, while retaining much of the notation and operations of evidence theory. We use the current standard notation throughout the paper.

\subsubsection{Information representation}\label{ir}
Consider an uncertain variable $X$ with frame of discernment (FoD) $\Omega = \{\omega_1, \cdots, \omega_n\}$. A belief function $Bel: 2^\Omega \rightarrow [0,1]$ satisfies $$\begin{aligned}
	\forall F_1, F_2, \dots, F_{k} \subseteq \Omega, Bel(F_1 \cup F_2 \cup \dots \cup F_k) \geq \sum_i Bel(F_i) - \sum_{i > j} Bel(F_i \cap F_j)+ \cdots + (-1)^{k+1} Bel(F_1 \cap F_2 \cap \dots \cap F_k),
\end{aligned}
$$ and represents information that restricts the possible value of $X$. Here $Bel(F_i)$ is the committed support for $X \in F_i$. Its dual, the plausibility function $Pl(F_i)=1-Bel(\overline{F_i})$, gives the possible support for the same proposition. Thus, $Bel$ and $Pl$ provide lower and upper support, respectively. An equivalent representation is the mass function $m$, or BPA, satisfying $m(F_i)\in[0,1]$ and $\sum_{F_i\subseteq \Omega}m(F_i)=1$. A set $F_i$ with $m(F_i)>0$ is a focal set, where $i$ is the decimal value of its binary membership code. The representations are related by the following invertible transforms \cite{smets2002application,denoeux2008conjunctive}:
$$
	\begin{aligned}
		& Bel(F_i)=\sum_{\emptyset\neq F_j\subseteq F_i}m(F_j), Bel(\emptyset)=0;m(F_i)=\sum_{F_j\subseteq F_i}(-1)^{|F_i|-|F_j|}Bel(F_j), m(\emptyset)=1-Bel(\Omega).\\
		& Pl(F_i)=\sum_{F_i\cap F_j\neq \emptyset}m(F_j), Pl(\emptyset)=0; m(F_i)=\sum_{F_j\subseteq F_i}(-1)^{|F_i|-|F_j|+1}Pl(\overline{F_j}), m(\emptyset)=1-Pl(\Omega).\\
	\end{aligned}$$
Two further representations are the implicability function $b$ and the commonality function $q$:
$$
	\begin{aligned}
		 b(F_i)=\sum_{F_j\subseteq F_i}m(F_j);~m(F_i)=\sum_{F_j\subseteq F_i}(-1)^{|F_i|-|F_j|}b(F_j), q(F_i)=\sum_{F_i\subseteq F_j}m(F_j);~m(F_i)=\sum_{F_i\subseteq F_j}(-1)^{|F_j|-|F_i|}q(F_j).
	\end{aligned}
	$$
For computation, write the BPA as $\boldsymbol{m}=[m(F_0),\cdots,m(F_{2^n-1})]^{T}$. The transforms above then have the matrix form \cite{smets2002application}:
	\begin{equation}\label{matrix_b_e}
		\begin{aligned}
			\boldsymbol{b}=\boldsymbol{m2b}\cdot \boldsymbol{m}, m2b(F_i,F_j)=
			\begin{cases}
				1 & F_j\subseteq F_i \\
				0 & \text{otherwise}\\
			\end{cases};\boldsymbol{q}=\boldsymbol{m2q}\cdot \boldsymbol{m}, m2q(F_i,F_j)=
			\begin{cases}
				1 & F_i\subseteq F_j \\
				0 & \text{otherwise}\\
			\end{cases}.
		\end{aligned}
	\end{equation}
\subsubsection{Credal level}

At the credal level, evidence sources are combined by transferring their belief masses. For independent and reliable sources, the conjunctive combination rule (CCR), also called unnormalized Dempster's rule, can be evaluated from either the BPA or the commonality function \cite{denoeux2008conjunctive}:
	\begin{equation}\label{ccr_e}
		\begin{aligned}
			m_{1\circledtiny{$\cap$}2}(F_i)
			=\sum_{F_j\cap F_k = F_i}m_1(F_j)m_2(F_k),q_{1\circledtiny{$\cap$}2}(F_i)&=q_1(F_i)q_2(F_i).
		\end{aligned}
	\end{equation}
Equation~(\ref{matrix_b_e}) also gives the matrix form $\boldsymbol{m}_{1\circledtiny{$\cap$}2}=\boldsymbol{S_{m_2}}\cdot \boldsymbol{m_1}$, where $\boldsymbol{S_{m_2}}=\boldsymbol{m2q}^{-1}\textbf{diag}(\boldsymbol{q_2})\boldsymbol{m2q}$. For independent sources when at least one source is reliable, the disjunctive combination rule (DCR) is defined by \cite{smets2002application}
	\begin{equation}
		\begin{aligned}
			m_{1\circledtiny{$\cup$}2}(F_i)=\sum_{F_j\cup F_k = F_i}m_1(F_j)m_2(F_k),~b_{1\circledtiny{$\cup$}2}(F_i)=b_1(F_i)b_2(F_i).
		\end{aligned}
	\end{equation}
The $\alpha$-junction generalizes these rules through a parameterized matrix construction \cite{smets2002application}:
	\begin{equation}\label{alpha_j_sum}
		\begin{aligned}
			\boldsymbol{m}_{1\circledtiny{$\cap$}^{\alpha}2}=\boldsymbol{K}^{\cap,\alpha}_{m_2}\cdot \boldsymbol{m_1},~\boldsymbol{m}_{1\circledtiny{$\cup$}^{\alpha}2}=\boldsymbol{K}^{\cup,\alpha}_{m_2}\cdot \boldsymbol{m_1}.
		\end{aligned}
	\end{equation}
For the conjunctive $\alpha$-junction $\circledsmall{$\cap$}^{\alpha}$, the neutral element is $m_{\Omega}$, so $m\circledsmall{$\cap$}^{\alpha}m_{\Omega}=m$. Its matrix $\boldsymbol{K}^{\cap,\alpha}_{m}$ is
	\begin{equation}\label{alpha_j_con}
		\begin{aligned}
			\boldsymbol{K}^{\cap,\alpha}_{m} = \sum_{F_i\subseteq\Omega}m(F_i)\cdot\boldsymbol{K}^{\cap,\alpha}_{F_i},~\boldsymbol{K}^{\cap,\alpha}_{F_i}=
			\begin{cases}
				\boldsymbol{I} & F_i = \Omega, \\
				\prod_{\omega\notin F_i}\boldsymbol{K}^{\cap,\alpha}_{\overline{\omega}} & F_i\subset \Omega,\\
			\end{cases},{K}^{\cap,\alpha}_{\overline{\omega}}(F_i,F_j)=\begin{cases}
				1 & \omega\notin F_i, F_j=F_i\cup\{\omega\},\\
				\alpha & \omega\notin F_j, F_i=F_j,\\
				1-\alpha & \omega\notin F_j, F_i=F_j\cup\{\omega\},\\
				0 & \text{otherwise},\\
			\end{cases}
		\end{aligned}
	\end{equation}
where $\alpha\in[0,1]$. The cases $\alpha=1$ and $\alpha=0$ give the CCR and the conjunctive exclusive combination rule (CECR), respectively:
	\begin{equation}
		\begin{aligned}
			m_1\circledsmall{$\underline{\cap}$}m_2(F_i)=\sum_{F_i=(F_j\cap F_k)\cup(\overline{F_j}\cap \overline{F_k})}m_1(F_j)m_2(F_k).
		\end{aligned}
	\end{equation}
For the disjunctive $\alpha$-junction $\circledsmall{$\cup$}^{\alpha}$, the neutral element is $m_{\emptyset}$, so $m\circledsmall{$\cup$}^{\alpha}m_{\emptyset}=m$. Its matrix $\boldsymbol{K}^{\cup,\alpha}_{m}$ is
	\begin{equation}\label{alpha_j_dis}
		\begin{aligned}
			\boldsymbol{K}^{\cup,\alpha}_{m} = \sum_{F_i\subseteq\Omega}m(F_i)\cdot\boldsymbol{K}^{\cup,\alpha}_{F_i},~\boldsymbol{K}^{\cup,\alpha}_{F_i}=
			\begin{cases}
				\boldsymbol{I} & F_i = \emptyset, \\
				\prod_{\omega\in F_i}\boldsymbol{K}^{\cup,\alpha}_{{\omega}} & F_i\in 2^\Omega\setminus\{\emptyset\},\\
			\end{cases},{K}^{\cup,\alpha}_{{\omega}}(F_i,F_j)=\begin{cases}
				1 & \omega\notin F_j, F_i=F_j\cup\{\omega\},\\
				\alpha & \omega\in F_j, F_i=F_j,\\
				1-\alpha & \omega\notin F_i, F_j=F_i\cup\{\omega\},\\
				0 & \text{otherwise},\\
			\end{cases}
		\end{aligned}
	\end{equation}
Here $\alpha=1$ gives the DCR, and $\alpha=0$ gives the disjunctive exclusive combination rule (DECR):
	\begin{equation}\label{decr}
		\begin{aligned}
			m_1\circledsmall{$\underline{\cup}$}m_2(F_i)=\sum_{F_i=(F_j\cap \overline{F_k})\cup(\overline{F_j}\cap F_k)}m_1(F_j)m_2(F_k).
		\end{aligned}	
	\end{equation}
The $\alpha$-junction is defined by these algebraic properties; its broader interpretation remains open.

\subsubsection{Pignistic level}

When no further evidence is available, a probability transformation allocates belief masses to singletons for decision making. The linear pignistic transformation is

	\begin{equation}\label{ppt_e}
		\begin{aligned}
			BetP_{m}(\omega)=\sum_{\omega\in F_i}\frac{m(F_i)}{(1-m(\emptyset))|F_i|}.
		\end{aligned}
	\end{equation}
Several alternative probability transformations have also been proposed \cite{han2016evaluation}. The plausibility transformation, based on Dempster's semantic consistency, is \cite{cobb2006plausibility,cui2023plausibility}
	\begin{equation}\label{ptm_e}
		\begin{aligned}
			\mathrm{PlP}_m(\omega_i)=\frac{pl_m(\omega_i)}{\sum_{\omega_j\in\Omega}pl_m(\omega_j)},
		\end{aligned}
	\end{equation}
where $pl_m(\omega)=Pl(\{\omega\})$ denotes the contour function.

\subsubsection{Operations on product space}

Product-space operations \cite{denoeux2006classification} support both the Generalized Bayesian Theorem (GBT) \cite{smets1993belief} and the Valuation-Based System (VBS) \cite{shenoy1990axioms}. Let $m^{\Omega \times \Theta}$ be a mass function on $\Omega \times \Theta$. Its marginalization to $\Omega$ is
	\begin{equation}\label{margin_e}
		\begin{aligned}
			 m^{\Omega \times \Theta\downarrow\Omega}(F_i)=\sum_{\{G_i\subseteq \Omega \times \Theta | {\rm{Proj}}_{\Omega}(G_i)=F_i\}}m^{\Omega \times \Theta}(G_i), {\rm{Proj}}_{\Omega}(G_i) = \{\omega\in\Omega| \exists \theta\in \Theta, (\omega,\theta)\in G_i\}.
		\end{aligned}
	\end{equation}
Vacuous extension maps a mass function $m^{\Omega}$ to $\Omega \times \Theta$:
	\begin{equation}\label{vacuous_e}
		\begin{aligned}
			m^{\Omega\uparrow\Omega\times\Theta}(G_i)=\begin{cases}
				m^{\Omega}(F_i) & F_i\subseteq \Omega,~G_i= F_i\times\Theta, \\
				0 & \text{otherwise}.\\
			\end{cases}
		\end{aligned}
	\end{equation}
Mass functions defined on different FoDs, such as $m^{\Omega}_1$ and $m^{\Theta}_2$, can then be extended, combined, and projected as
$m^{\Omega}_{1\circledtiny{$\cdot$}2} = (m^{\Omega\times\Theta}_1 \circledsmall{$\cdot$} m^{\Omega\times\Theta}_2)^{\downarrow\Omega}$; $m^{\Theta}_{1\circledtiny{$\cdot$}2} =( m^{\Omega\times\Theta}_1 \circledsmall{$\cdot$} m^{\Omega\times\Theta}_2)^{\downarrow\Theta}$.
Combining with a categorical mass function, which has a single focal set, defines conditioning. Given $H_i\subseteq\Theta$, the conditional mass on $\Omega$ is $m^\Omega[H_i]=(m^{\Omega\times\Theta}\circledsmall{$\cap$}m^{\Omega\times\Theta}_{H_i})^{\downarrow\Omega}$, where $m_{H_i}\equiv\{m(H_i)=1\}$.
Ballooning extension maps the conditional mass $m^\Omega[H_i]$ back to $\Omega \times \Theta$:
	\begin{equation}\label{ballooning_e}
		\begin{aligned}
			m^{\Omega}[H_i]^{\Uparrow \Omega \times \Theta}(G_i) = 
			\begin{cases}
				m^{\Omega}[H_i](F_i) &  G_i = (F_i \times H_i) \cup (\Omega \times (\Theta \setminus H_i)), F_i \subseteq \Omega, \\
				0 & \text{otherwise}.
			\end{cases}
		\end{aligned}
	\end{equation}
Section~\ref{method3} implements these product-space operations on quantum circuits.

\subsection{Quantum computing}

Quantum computing processes information through quantum-state evolution and measurement \cite{Biamonte2017quantum}. This subsection reviews the concepts used in the circuit constructions.

\subsubsection{Quantum state}\label{qs}

An isolated quantum system is described by a state vector $\ket{\psi}$ in a complex Hilbert space. The computational basis states of one qubit are $\ket{0}$ and $\ket{1}$. An $n$-qubit state is a superposition of all $2^n$ computational basis states:
$$\ket{\psi}\equiv a_0\ket{0}^n+a_1\ket{0}^{n-1}\ket{1}+\cdots+a_{2^n-1}\ket{1}^n=\begin{bmatrix}
	a_0\\
	\vdots\\
	a_{2^n-1}
\end{bmatrix},$$
where $|a_0|^2+|a_1|^2+\cdots+|a_{2^n-1}|^2=1$. Multiple states form a joint system through the tensor product, written as $\ket{\psi_0}\otimes\ket{\psi_1}\otimes\cdots\otimes\ket{\psi_k}=\ket{\psi_0\psi_1\cdots\psi_k}$.

\subsubsection{Quantum evolution}\label{qe}
The evolution of a closed quantum system is governed by a unitary operator $\boldsymbol{U}$. If $\ket{\psi}$ evolves to $\ket{\psi'}$ from $t$ to $t'$, then $\ket{\psi'}=\boldsymbol{U}_{t\rightarrow t'}\ket{\psi}$. Unitarity makes the evolution deterministic and reversible. Quantum gates implement these operators. Table \ref{quantum_gate_t} lists the gates used below.

\begin{table}[H]
	\centering
	\caption{Quantum gates used in this paper, where $\boldsymbol{I}_n$ is the $n$-dimensional identity matrix.}
	\label{quantum_gate_t}
	\resizebox{0.75\linewidth}{!}{
		\begin{tabular}{ccccc}
			\Xhline{1.5pt}
			X gate & RY gate & C-NOT gate & Toffoli gate & Control-RY gate\\
			\hline
			$\Qcircuit @C=.3em @R=.4em { & \gate{X} & \qw }$ &
			$\Qcircuit @C=.3em @R=.4em { & \gate{R_\mathrm{Y}(\theta)} & \qw }$ &
			$\Qcircuit @C=.3em @R=.4em { \lstick{c}& \ctrl{1} &  \qw \\ \lstick{t}& \targ &  \qw }$ &
			$\Qcircuit @C=.3em @R=.4em { \lstick{c}& \ctrl{1} &  \qw \\ \lstick{c}& \ctrl{1} &  \qw \\ \lstick{t}& \targ &  \qw }$ &
			$\Qcircuit @C=.3em @R=.4em { \lstick{c}& \ctrl{1} &  \qw \\ \lstick{t}& \gate{R_\mathrm{Y}(\theta)} &  \qw }$ \\
			\hline
			$\begin{bmatrix} 0 & 1 \\ 1 & 0\end{bmatrix}$ &
			$\begin{bmatrix} \cos\!\tfrac{\theta}{2} &  -\sin\!\tfrac{\theta}{2} \\  \sin\!\tfrac{\theta}{2} &  \cos\!\tfrac{\theta}{2}\end{bmatrix}$ &
			$\begin{bmatrix}\boldsymbol{I}_2 & 0 \\ 0 & \boldsymbol{X}\end{bmatrix}$ &
			$\begin{bmatrix}\boldsymbol{I}_6 & 0 \\ 0 & \boldsymbol{X}\end{bmatrix}$ &
			$\begin{bmatrix}\boldsymbol{I}_2 & 0 \\ 0 & \boldsymbol{R}_{\rm{Y}}(\theta)\end{bmatrix}$ \\
			\Xhline{1.5pt}
	\end{tabular}}
\end{table}

\subsubsection{Quantum measurement}\label{qm}

A physical quantity is measured through its Hermitian operator. Measurement returns an eigenvalue and projects the system onto the corresponding eigenstate. Quantum circuits commonly use computational-basis measurements. For a superposition, the probability of a basis outcome equals the squared modulus of its amplitude. For example, measuring $\ket{\psi}=a\ket{0}+b\ket{1}$ returns $\ket{1}$ with probability $|b|^2$.

\subsubsection{Quantum entanglement}\label{qen}

Entanglement is a nonseparable dependence among quantum subsystems. A separable joint state factors into subsystem states; an entangled state does not. For example, $\ket{\Phi^+}=a\ket{00}+b\ket{11}$ is entangled when both coefficients are nonzero. Controlled gates can generate such dependencies.

The constructions below use these four properties: an $n$-qubit state spans $2^n$ basis states, unitary operators update amplitudes and phases, projective measurements extract probabilities, and controlled gates create dependencies between qubits.

\section{Representing and implementing belief functions on quantum circuits}
\label{method1}

\subsection{Motivation}

A mass function on an $n$-element FoD has $2^n$ coordinates with the same Boolean-lattice structure as the computational basis of an $n$-qubit register. Each focal set is identified by its membership vector, with one qubit storing the membership of each frame element. A computational-basis query on the register therefore evaluates a set-theoretic property of the focal set.

\begin{figure}[H]
	\centering
	\includegraphics[width=0.75\textwidth]{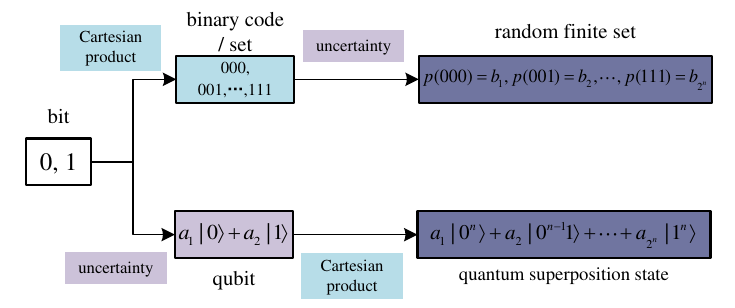}
	\caption{Correspondence between mass functions and quantum computational-basis states.}
	\label{quantum_classic_r}
\end{figure}

A singleton probability representation evaluates a set proposition $A\subseteq\Omega$ by aggregating elementary events. Element-qubit encoding addresses $A$ directly as a basis state or a control condition. Controlled gates can then test predicates such as $A\subseteq F$ and $\omega_i\in F$. Figure \ref{quantum_classic_r} relates bit strings to random finite sets under this encoding.

\subsection{Encoding mass function as superposition state}

The encoding maps FoD elements to qubits, focal sets to computational-basis states, a mass function to a quantum state, and belief masses to measurement probabilities.
\begin{definition}
	Consider a mass function $m$ on the FoD $\Omega=\{\omega_1,\dots,\omega_n\}$. For any focal set $F\subseteq\Omega$, define its membership vector by
	\[
	\chi(F)=(\chi_1(F),\dots,\chi_n(F)),\qquad
	\chi_j(F)=
	\begin{cases}
	1, & \omega_j\in F,\\
	0, & \omega_j\notin F.
	\end{cases}
	\]
	The associated computational-basis state is denoted by
	\[
	\ket{F}:=\ket{\chi_1(F)\cdots\chi_n(F)} .
	\]
	An $n$-qubit mass function quantum state (MFQS) is
	\begin{equation}\label{q_bpa}
		\ket{m}=\sum_{F\subseteq\Omega}\sqrt{m(F)}\ket{F}.
	\end{equation}
	A computational-basis measurement returns the focal set $F$ with probability $m(F)$.
\end{definition}

\begin{remark}
	Orthogonality of the states $\ket{F}$ serves as a computational encoding and does not make the corresponding focal sets mutually exclusive propositions. A measurement samples one focal set from the random set represented by $m$. The bit pattern of $\ket{F}$ preserves its membership information, including the semantics of non-singleton focal sets. Controlled operations can therefore evaluate set predicates directly.
\end{remark}

\begin{remark}
	We use nonnegative real amplitudes, so squared amplitudes recover the belief masses. Phase carries no additional uncertainty in the present model. Complex-valued evidence theory \cite{xiao2023complex} and phase-based extensions lie outside the current scope.
\end{remark}

\subsection{Implementation of mass function quantum state}

Preparing a general MFQS is an amplitude-loading problem. The construction in \cite{zhou2023bf} can be expressed as uniformly controlled $R_Y$ rotations. Let $m$ be a mass function on $\Omega$, and initialize an $n$-qubit register in $\ket{0}^{\otimes n}$.

\begin{itemize}
	\item \textbf{Initialization.}  
	Apply $X^{\otimes n}$ and obtain
	\[
	\ket{\psi_1}=X^{\otimes n}\ket{0}^{\otimes n}=\ket{1}^{\otimes n}.
	\]
	
	\item \textbf{Uniformly controlled rotations.}  
	For each qubit $q_k$ ($k=0:n-1$), apply the controlled operation
	\[
	U_k=\sum_{t=0}^{2^k-1}\bigl(\ket{t}\!\bra{t}\bigr)_{0:k-1}\otimes R_Y(\theta_{k,t})_k .
	\]  
	After acting on $\ket{\psi_k}$, the system evolves as
	$\ket{\psi_{k+1}}=U_k\ket{\psi_k}$,
	where $t$ denotes the computational branch determined by the first $k$ qubits.  
	
	\item \textbf{Angle selection.}  
	For each branch $t$, define the partial sums
	\[
	S^{(1)}_{k,t}=\!\!\sum_{\substack{F:\mathrm{prefix}(F,k)=t,\omega_{k+1}\in F}} m(F),\quad
	S^{(0)}_{k,t}=\!\!\sum_{\substack{F:\mathrm{prefix}(F,k)=t,\omega_{k+1}\notin F}} m(F).
	\]  
	The rotation angle is chosen by $\tan^2\!\Big(\tfrac{\theta_{k,t}}{2}\Big)=\frac{S^{(0)}_{k,t}}{S^{(1)}_{k,t}}$,
	ensuring that along branch $t$, the measurement probability ratio matches
	\[
	\Pr[\omega_{k+1}\notin F]:\Pr[\omega_{k+1}\in F]=S^{(0)}_{k,t}:S^{(1)}_{k,t}.
	\]
	
\item \textbf{Final state.}  
	After all controlled rotations,
	\[
	\ket{\psi_{\mathrm{final}}}
	=\Big(\prod_{k=0}^{n-1}U_k\Big)\,X^{\otimes n}\ket{0^n},
	\]
	where $|\braket{F|\psi_{\mathrm{final}}}|^2 = m(F)$ for every $F\subseteq\Omega$.
\end{itemize}

Figure \ref{s3f2} shows the circuit for a three-element FoD. The construction encodes an arbitrary mass function and requires exponentially many controlled rotations in the worst case. Accordingly, the resource analysis in later sections assumes prepared MFQS inputs and focuses on circuit-level focal-set operations. Structured mass functions may also admit simpler preparation.

\begin{figure}[htbp!]
	\centering
	\begin{tikzpicture}[
		x=0.92cm,y=0.75cm,
		wire/.style={draw=qWire,line width=0.55pt},
		xgate/.style={draw=qXDraw,fill=qXFill,rounded corners=2pt,minimum width=0.55cm,minimum height=0.42cm,font=\small},
		rygate/.style={draw=qRyDraw,fill=qRyFill,rounded corners=2pt,minimum width=1.15cm,minimum height=0.42cm,font=\scriptsize},
		ctrl/.style={circle,fill=qCtrl,inner sep=1.5pt},
		octrl/.style={circle,draw=qCtrl,fill=white,line width=0.65pt,inner sep=1.5pt},
		ann/.style={font=\scriptsize,color=qWire}
	]
	\foreach \y/\q in {0/q_0,-1/q_1,-2/q_2}{
		\draw[wire] (0,\y)--(13.0,\y);
		\node[left,font=\small] at (0,\y) {$\q:\ket{0}$};
		\node[xgate] at (0.75,\y) {$X$};
	}
	\node[rygate] at (1.85,0) {$R_Y(\theta_0)$};
	\foreach \x/\cstyle/\lab in {3.2/octrl/$R_Y(\theta_{1,0})$,4.9/ctrl/$R_Y(\theta_{1,1})$}{
		\draw[wire] (\x,0)--(\x,-1);
		\node[\cstyle] at (\x,0) {};
		\node[rygate] at (\x,-1) {\lab};
	}
	\foreach \x/\ca/\cb/\lab in {
		6.6/octrl/octrl/$R_Y(\theta_{2,0})$,
		8.3/ctrl/octrl/$R_Y(\theta_{2,1})$,
		10.0/octrl/ctrl/$R_Y(\theta_{2,2})$,
		11.7/ctrl/ctrl/$R_Y(\theta_{2,3})$}{
		\draw[wire] (\x,0)--(\x,-2);
		\node[\ca] at (\x,0) {};
		\node[\cb] at (\x,-1) {};
		\node[rygate] at (\x,-2) {\lab};
	}
	\node[ann,above] at (1.25,0.45) {flip};
	\node[ann,above] at (8.3,0.45) {branch-conditioned rotations};
	\end{tikzpicture}
	\caption{Uniformly controlled $R_Y$ preparation of a general MFQS in a $3$-qubit system.}
	\label{s3f2}
\end{figure}

The poss-transferable mass function is a structured class with a simpler state preparation. It is obtained through an invertible transformation from a possibility distribution to a belief function \cite{zhou2024cd}.
\begin{definition}
	Consider a possibility distribution $\pi$ on an $n$-element FoD. The poss-transferable mass function is defined as \cite{zhou2024cd}
		\begin{equation}
			m_{Poss}(F_i)=\prod_{\omega\in F_i}\pi(\omega)\prod_{\omega\notin F_i}(1-\pi(\omega))=\circledlarge{$\cap$}_{\omega\in\Omega}\{\Omega\setminus\{\omega\}\}^{\pi(\omega)},
		\end{equation}
	where $F_{i}^{\sigma}\equiv\{m(F_i)=1-\sigma,~m(\Omega)=\sigma\}$ denotes a simple mass function. The transformation inverts canonical decomposition. A mass function is poss-transferable when its canonical decomposition yields a possibility distribution.
\end{definition}

\begin{theorem}
	For an $n$-element FoD, the quantum state of a poss-transferable mass function can be implemented using exactly $n$ single-qubit $R_Y$ gates, with rotation parameters determined by its contour function.
\end{theorem}

\begin{proof}
	Let $m_{Poss}$ be a poss-transferable mass function and write $\pi_j=\pi(\omega_j)$. Its contour value at $\omega_j$ is $\pi_j$: summing $m_{Poss}(F)$ over all focal sets that contain $\omega_j$ marginalizes the product distribution and gives $\Pr(\omega_j\in F)=\pi_j$.
	
	Prepare each qubit independently from $\ket{0}$ by applying $R_Y(\theta_j)$ to $q_{j-1}$, where
	\[
		\sin^2\frac{\theta_j}{2}=\pi_j,\qquad
		\cos^2\frac{\theta_j}{2}=1-\pi_j .
	\]
	Equivalently, for $0<\pi_j<1$, $\theta_j=2\arctan\sqrt{\pi_j/(1-\pi_j)}$, with the boundary cases understood by continuity. The resulting product state is
	\[
	\bigotimes_{j=1}^{n}\left(\sqrt{1-\pi_j}\ket{0}+\sqrt{\pi_j}\ket{1}\right).
	\]
	Expanding it in the focal-set basis yields
	\[
	\sum_{F\subseteq\Omega}
	\sqrt{\prod_{\omega_j\in F}\pi_j
	\prod_{\omega_j\notin F}(1-\pi_j)}\ket{F}
	=
	\sum_{F\subseteq\Omega}\sqrt{m_{Poss}(F)}\ket{F}
	=\ket{m_{Poss}}.
	\]
	No entangling gate is required, so the construction uses exactly $n$ single-qubit $R_Y$ gates.
\end{proof}

\begin{remark}\label{r3}
	Poss-transferable mass functions yield separable MFQSs. A general mass function may require entanglement when its focal-set weights do not factor into independent element-wise memberships.
\end{remark}

\begin{remark}
	Poss-transferable and consonant mass functions both admit unique possibility-distribution transforms, but they have different constructions and properties \cite{zhou2024cd}.
\end{remark}

\subsection{Implementation of belief functions on quantum circuits}

Section \ref{ir} defines equivalent mass-function representations through set-inclusion relations. Under MFQS encoding, these relations become qubit predicates. Controlled gates and ancilla measurements can therefore estimate commonality, implicability, and singleton contours.

\begin{definition}
	Let $\ket{m}$ be a MFQS encoded on data qubits $q_0,\dots,q_{n-1}$, and let $a$ be an ancilla initialized in $\ket{0}$. For a query set $A\subseteq\Omega$, define two reversible predicates:
	\[
	I_q(F;A)=
	\begin{cases}
	1, & A\subseteq F,\\
	0, & \text{otherwise},
	\end{cases}
	\qquad
	I_b(F;A)=
	\begin{cases}
	1, & F\subseteq A,\\
	0, & \text{otherwise}.
	\end{cases}
	\]
	The corresponding controlled gates act as
	\[
	\begin{aligned}
	U_q(A):\ket{F}\ket{a}\mapsto
	\ket{F}\ket{a\oplus I_q(F;A)},U_b(A):\ket{F}\ket{a}\mapsto
	\ket{F}\ket{a\oplus I_b(F;A)}.
	\end{aligned}
	\]
	When they are applied to $\ket{m}\ket{0}$, the states become
	\[
	U_q(A)\ket{m}\ket{0}
	=
	\sum_{F\subseteq\Omega}\sqrt{m(F)}\ket{F}\ket{I_q(F;A)}, U_b(A)\ket{m}\ket{0}
	=
	\sum_{F\subseteq\Omega}\sqrt{m(F)}\ket{F}\ket{I_b(F;A)}.
	\]
	Therefore, measuring the ancilla gives
	\[
	\begin{aligned}
	\Pr[a=1\mid U_q(A)] = \sum_{A\subseteq F}m(F)=q(A),\Pr[a=1\mid U_b(A)] = \sum_{F\subseteq A}m(F)=b(A).
	\end{aligned}
	\]
	The specific readout circuits are shown in Figure \ref{quantum_belief_functions}.
	\begin{figure}[htbp!]
		\centering
		\begin{tikzpicture}[
			x=0.74cm,y=0.68cm,
			wire/.style={draw=qWire,line width=0.55pt},
			prep/.style={draw=qOpDraw,fill=qOpFill,rounded corners=2pt,minimum width=0.72cm,minimum height=1.12cm,font=\small},
			mcx/.style={draw=qXDraw,fill=qXFill,rounded corners=2pt,minimum width=0.68cm,minimum height=0.38cm,font=\scriptsize},
			meas/.style={draw=qMeasDraw,fill=qMeasFill,rounded corners=2pt,minimum width=0.72cm,minimum height=0.38cm,font=\scriptsize},
			ctrl/.style={circle,fill=qCtrl,inner sep=1.45pt},
			octrl/.style={circle,draw=qCtrl,fill=white,line width=0.65pt,inner sep=1.45pt},
			txt/.style={font=\tiny,color=qWire}
		]
		\begin{scope}
			\node[font=\scriptsize\bfseries,anchor=west] at (0,0.85) {(a) Commonality: $A\subseteq F$};
			\foreach \y/\q in {0/q_0,-0.58/q_1,-1.16/q_{n-1}}{
				\draw[wire] (0,\y)--(6.9,\y);
				\node[left,font=\scriptsize] at (0,\y) {$\q$};
			}
			\node[left,font=\scriptsize] at (0,-0.87) {$\vdots$};
			\draw[wire] (0,-1.86)--(4.75,-1.86);
			\node[left,font=\scriptsize] at (0,-1.86) {$a_q:\ket{0}$};
			\node[prep] at (0.72,-0.58) {$U_m$};
			\draw[wire] (2.35,0)--(2.35,-1.86);
			\node[ctrl] at (2.35,0) {};
			\node[ctrl] at (2.35,-0.58) {};
			\node[txt] at (2.35,-0.92) {$\vdots$};
			\node[mcx] at (2.35,-1.86) {$X$};
			\node[txt,anchor=west] at (2.62,0.28) {controls in $A$};
			\node[meas] at (4.05,-1.86) {meas.};
			\node[txt] at (5.65,-2.85) {$\Pr(a_q=1)=q(A)$};
		\end{scope}
		
		\begin{scope}[shift={(8.25,0)}]
			\node[font=\scriptsize\bfseries,anchor=west] at (0,0.85) {(b) Implicability: $F\subseteq A$};
			\foreach \y/\q in {0/q_0,-0.58/q_1,-1.16/q_{n-1}}{
				\draw[wire] (0,\y)--(6.9,\y);
				\node[left,font=\scriptsize] at (0,\y) {$\q$};
			}
			\node[left,font=\scriptsize] at (0,-0.87) {$\vdots$};
			\draw[wire] (0,-1.86)--(4.75,-1.86);
			\node[left,font=\scriptsize] at (0,-1.86) {$a_b:\ket{0}$};
			\node[prep] at (0.72,-0.58) {$U_m$};
			\draw[wire] (2.35,0)--(2.35,-1.86);
			\node[octrl] at (2.35,0) {};
			\node[octrl] at (2.35,-0.58) {};
			\node[txt] at (2.35,-0.92) {$\vdots$};
			\node[mcx] at (2.35,-1.86) {$X$};
			\node[txt,anchor=west] at (2.62,0.28) {controls outside $A$};
			\node[meas] at (4.05,-1.86) {meas.};
			\node[txt] at (5.65,-2.85) {$\Pr(a_b=1)=b(A)$};
		\end{scope}
		\end{tikzpicture}
		\caption{Unified ancilla readout of belief function representations from an MFQS.}
		\label{quantum_belief_functions}
	\end{figure}
\end{definition}

These circuits implement inclusion-based belief representations as reversible predicates. The data register retains the MFQS, and the ancilla records whether a sampled focal set satisfies the query. Element-qubit controls thus act directly on membership information before any reduction to singleton probabilities.

\subsection{Discussion}
MFQSs retain focal sets before pignistic reduction: each data qubit records the membership of one frame element, and controlled gates evaluate predicates on the resulting bit pattern. General amplitude loading can be expensive. Poss-transferable mass functions form a separable class that requires only single-qubit rotations. Once an MFQS is prepared, controlled gates and ancilla measurements provide commonality, implicability, and contour values.

\section{Credal level on quantum circuits}
\label{method2}

\subsection{Combination rules for independent sources on quantum circuits}

The CCR, DCR, CECR, and DECR combine source weights by multiplication and focal sets by Boolean operations. Quantum-state composition multiplies amplitudes, and controlled gates evaluate Boolean functions. These two properties support a unified circuit construction for independent-source combination rules.

\subsubsection{Boolean algebra-based combination rule}

\begin{definition}\label{bacr_e}
	For $k$ mass functions $m_1,\cdots, m_k$ from independent sources, the Boolean algebra-based combination rule (BACR) is
	\begin{equation}
		m_{k + 1}(F_{{i}_{k+1}})=\sum_{\mathfrak{B}(F_{i_1},\cdots,F_{i_k})=F_{i_{k+1}}}\prod_{j=1}^{k}m_j(F_{i_j}),
	\end{equation}
	where $m_{k + 1}$ is the combined mass function, $F_{i_j}$ is a focal set of source $j$\footnote{The index $i_j$ is the decimal value of the membership code for a focal set from source $j$.}, and $\mathfrak{B}(F_{i_1},\cdots,F_{i_k})$ is a Boolean operation on the input focal sets.
\end{definition}

In Definition~\ref{bacr_e}, intersection $\mathfrak{B}(F_{i_1},\ldots,F_{i_k})=F_{i_1}\cap\cdots\cap F_{i_k}$ gives the CCR, and union gives the DCR. The choice $\mathfrak{B}(F_{i_1},\ldots,F_{i_k})=(F_{i_1}\cap F_{i_2})\cup\cdots\cup(F_{i_{k-1}}\cap F_{i_k})$ gives the $(K\!-\!1)$-out-of-$K$ rule~\cite{pichon2015consistency}. BACR therefore unifies independent-source rules defined by Boolean focal-set operations. For categorical inputs, it reduces to the corresponding classical Boolean operation.
\begin{itemize}
	\item \textbf{$t$-layer Boolean algebra operation:} The integer $t$ is the minimum circuit depth required to evaluate $\mathfrak{B}$ when independent operations run in parallel. For example, in $\mathfrak{B}=(\{\omega_1\omega_2\}\cap\{\omega_2\omega_3\})\cup (\{\omega_2\omega_4\}\cap\{\omega_4\omega_5\})$, the two intersections form the first layer and their union forms the second. Thus, $\mathfrak{B}$ and its BACR are two-layer operations.
	\item \textbf{Operation component:} An operation component is a maximal Boolean operation within one layer that maps its inputs to one intermediate result. In the example above, the two intersections are separate operation components.
\end{itemize}

\subsubsection{BACR on quantum circuits}

Existing quantum implementations of CCR and DCR often translate matrix equations into general linear-system or variational solvers such as HHL and VQLS \cite{zhou2023bf,luo2024variational}. MFQS encoding permits a more direct construction. A focal set is a membership bit string, so controlled gates can evaluate its set operations before measurement. We implement credal-level fusion as Boolean transformations of focal-set registers.

\begin{definition}\label{bacr_qc_d}
	Let $\{\ket{m_j}\}_{j=1}^k$ be $k$ MFQSs, each encoded in an $n$-qubit system 
	$q_{0_j},\dots,q_{(n-1)_j}$, where $q_{i_j}$ corresponds to the basic element $\omega_{i+1}$ in $\ket{m_j}$.  
	The output state $\ket{m_{k+1}}$ is encoded on $q_{0_{k+1}},\dots,q_{(n-1)_{k+1}}$ and represents the BACR result.
	For a single operation component in $\mathfrak{B}$, the state evolution is as follows.
	
	\medskip
	\noindent\textbf{(a) Negation.}  
	For the complement of $\ket{m_j}$, apply $X^{\otimes n}$:
	\[
	\begin{aligned}
		X^{\otimes n}\ket{m_j}
		&= X^{\otimes n}\!\!\sum_{F_i\subseteq \Omega}\sqrt{m(F_i)}\ket{\mathrm{bin}(i)} \longrightarrow \sum_{F_i\subseteq \Omega}\sqrt{m(\overline{F_i})}\ket{\mathrm{bin}(i)}
		= \ket{\overline{m}_j}.
	\end{aligned}
	\]
	
	\medskip
	\noindent\textbf{(b) Intersection (multi-source CCR).}  
	For the conjunctive combination of $\ket{m_1},\dots,\ket{m_k}$,  
	the $i$-th output qubit is obtained via a multi-controlled NOT:
	\[
	\begin{aligned}
		&\ket{\psi_0} = \ket{m_1}\cdots\ket{m_k}\ket{0}, \ket{\psi_1} = C^{\{q_{i_1},\dots,q_{i_k}\}}X(q_{i_{k+1}})\ket{\psi_0} \rightarrow \ket{m_1}\cdots\ket{m_k}\Bigl(
		\sqrt{\sum_{F:\,\omega_{i+1}\in \cap_{j=1}^k F_j}}\,\ket{1}
		+ \sqrt{\sum_{F:\,\omega_{i+1}\notin \cap_{j=1}^k F_j}}\,\ket{0}
		\Bigr).
	\end{aligned}
	\]
	
	\medskip
	\noindent\textbf{(c) Union (multi-source DCR).}\par
	For the disjunctive combination of $\ket{m_1},\dots,\ket{m_k}$,
	the $i$-th output qubit is obtained via a negatively controlled NOT, followed by a correction $X$:
	\[
	\begin{aligned}
		\ket{\psi_0}&=\ket{m_1}\cdots\ket{m_k}\ket{0},\\
		\ket{\psi_1}&=
		C^{\{\overline{q_{i_1}},\dots,\overline{q_{i_k}}\}}
		X(q_{i_{k+1}})\ket{\psi_0}\longrightarrow \ket{m_1}\cdots\ket{m_k}\Bigl(
		\sqrt{\sum_{F:\,\omega_{i+1}\in \cup_{j=1}^k F_j}}\,\ket{1}
		+ \sqrt{\sum_{F:\,\omega_{i+1}\notin \cup_{j=1}^k F_j}}\,\ket{0}
		\Bigr), \\[0.8em]
		\ket{\psi_2}&=X(q_{i_{k+1}})\ket{\psi_1}\longrightarrow \ket{m_1}\cdots\ket{m_k}\Bigl(
		\sqrt{\sum_{F:\,\omega_{i+1}\in \cup_{j=1}^k F_j}}\,\ket{1}
		+ \sqrt{\sum_{F:\,\omega_{i+1}\notin \cup_{j=1}^k F_j}}\,\ket{0}
		\Bigr).
	\end{aligned}
	\]
	
	\medskip
	In the general BACR with multiple operation components and hierarchical layers,  
	the output of each component is written to $n$ ancilla qubits, which then serve as the input for the next layer.
\end{definition}

\begin{example}\label{e1}
	Consider the two mass functions
	\[
		m_1=\{0.2,0.1,0.4,0.3\},\qquad
		m_2=\{0.05,0.13,0.02,0.8\},
	\]
	and the Boolean operation
	$\mathfrak{B}=(F_{i_1}\cap F_{i_2})\cup
	(\overline{F_{i_1}}\cap\overline{F_{i_2}})$.
	Definition~\ref{bacr_e} gives
	$m_3=\{0.229,0.143,0.357,0.271\}$.
	Following Definition~\ref{bacr_qc_d}, Figure~\ref{tbm_qc_f1} shows the layered circuit that evaluates the two operation components and then writes their union to the output register. The measured probability vector of $q_{0_3}q_{1_3}$ is
	$[0.230,0.143,0.356,0.271]^T$.
	\begin{figure}[htbp!]
		\centering
		\includegraphics[width=0.85\textwidth]{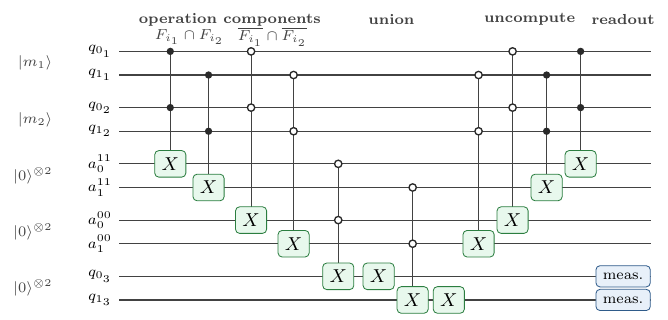}
		\caption{Layered BACR circuit for Example~\ref{e1}.}
		\label{tbm_qc_f1}
	\end{figure}
\end{example}

\begin{theorem}
	For a Boolean algebra operation $\mathfrak{B}$, the squared amplitude of $\ket{m_{k+1}}$ constructed in Definition \ref{bacr_qc_d} coincides with the combined mass function $m_{k+1}$ in Definition \ref{bacr_e}.
\end{theorem}

\begin{proof}
	Consider one layer of Boolean components; composing these layers gives a general BACR. Prepare
	\[
	\ket{\psi_0}
	= \sum_{F_{i_1},\dots,F_{i_k}\subseteq\Omega}
	\Bigl(\prod_{j=1}^{k}\sqrt{m_j(F_{i_j})}\Bigr)
	\ket{F_{i_1}}\cdots\ket{F_{i_k}}\ket{0}^n .
	\]
	The controlled gates in Definition~\ref{bacr_qc_d} evaluate the Boolean predicate element by element, hence
	\[
	\ket{\psi_{\mathrm{out}}}
	= \sum_{F_{i_1},\dots,F_{i_k}\subseteq\Omega}
	\Bigl(\prod_{j=1}^{k}\sqrt{m_j(F_{i_j})}\Bigr)
	\ket{F_{i_1}}\cdots\ket{F_{i_k}}
	\ket{\mathfrak{B}(F_{i_1},\dots,F_{i_k})}.
	\]
	For a fixed output focal set $F_{i_{k+1}}$, define
	\[
		\mathcal{S}_{F_{i_{k+1}}}
		=\{(F_{i_1},\dots,F_{i_k}):
		\mathfrak{B}(F_{i_1},\dots,F_{i_k})=F_{i_{k+1}}\}.
	\]
	All branches in $\mathcal{S}_{F_{i_{k+1}}}$ contribute to the same measured outcome, whose probability is
	\[
	\sum_{(F_{i_1},\dots,F_{i_k})\in\mathcal{S}_{F_{i_{k+1}}}}
	\prod_{j=1}^{k}m_j(F_{i_j}),
	\]
	which is exactly Definition~\ref{bacr_e}.
\end{proof}

\begin{theorem}\label{th2+1}
	For two sources, CECR and DECR admit direct circuits without extra work qubits.
\end{theorem}

\begin{proof}
	For each element $\omega_j$, CECR and DECR only require the equivalence and exclusive-or of the two membership bits $q_{j_1}$ and $q_{j_2}$.  Starting from
	\[
	\ket{\psi_0}=\sum_{F_1,F_2\subseteq\Omega}\sqrt{m_1(F_1)m_2(F_2)}
	\ket{F_1}\ket{F_2}\ket{0}^n,
	\]
	apply CNOT gates from $q_{j_1}$ and $q_{j_2}$ to the target qubit $q_{j_3}$ for all $j$. The target register then stores the bitwise exclusive-or
	\[
	F_3=(F_1\cap\overline{F_2})\cup(\overline{F_1}\cap F_2),
	\]
	so its squared amplitudes reproduce DECR. Applying $X^{\otimes n}$ to the target register changes the exclusive-or into bitwise equivalence:
	\[
	F_3=(F_1\cap F_2)\cup(\overline{F_1}\cap\overline{F_2}),
	\]
	which gives CECR.  Since both constructions write directly to the target register using only CNOT and final $X$ gates, no additional ancilla register is needed.
\end{proof}
BACR gives belief function fusion a circuit semantics. Reversible Boolean predicates implement focal-set operations, and amplitudes retain the uncertainty weights.

\subsection{\texorpdfstring{$\alpha$}{alpha}-junction on quantum circuits}

The $\alpha$-junction is a parameterized combination rule derived from matrix calculus \cite{smets2002application}. It interpolates between boundary rules and requires a family of focal-set-indexed matrices in its classical form. MFQS encoding decomposes the calculation into an $\alpha$-dependent local transformation and controlled-gate extraction of focal sets.

According to Eqs.~(\ref{alpha_j_sum})-(\ref{alpha_j_dis}),  
the $\alpha$-junction can be written in a unified form as
\[
m_1 \circledsmall{$\cdot$}^\alpha m_2
= \sum_{F_i\subseteq\Omega} m_1(F_i)\,
\boldsymbol{K}^{\cdot,\alpha}_{F_i}\,\boldsymbol{m}_2
= \sum_{F_i\subseteq\Omega} m_1(F_i)\,
\boldsymbol{m}^{\cdot,\alpha}_{2,F_i},
\]
where $\cdot \in \{\cap,\cup\}$.  
Its main computational costs arise from constructing
$\boldsymbol{K}^{\cdot,\alpha}_{F_i}$ and $\boldsymbol{m}^{\cdot,\alpha}_{2,F_i}$.

\begin{theorem}\label{theorem_alpha}
	The operator $\boldsymbol{K}^{\cdot,\alpha}_{F_i}$ can be realized as a Kronecker product of $2$-dimensional matrices.  
	
	In the conjunctive case, the characteristic matrix is
	\[
	\boldsymbol{C}^{\cap,\alpha}=
	\begin{bmatrix}
		\alpha & 1 \\[0.3em]
		1-\alpha & 0
	\end{bmatrix},
	\boldsymbol{K}^{\cap,\alpha}_{F_i}
	=\bigotimes_{j=1}^{|\Omega|}
	\begin{cases}
		\boldsymbol{C}^{\cap,\alpha}, & \omega_{|\Omega|-j+1}\notin F_i, \\[0.3em]
		\boldsymbol{I}_2, & \omega_{|\Omega|-j+1}\in F_i .
	\end{cases}
	\]
	
	In the disjunctive case, the characteristic matrix is
	\[
	\boldsymbol{C}^{\cup,\alpha}=
	\begin{bmatrix}
		0 & 1-\alpha \\[0.3em]
		1 & \alpha
	\end{bmatrix},
	\boldsymbol{K}^{\cup,\alpha}_{F_i}
	=\bigotimes_{j=1}^{|\Omega|}
	\begin{cases}
		\boldsymbol{C}^{\cup,\alpha}, & \omega_{|\Omega|-j+1}\in F_i, \\[0.3em]
		\boldsymbol{I}_2, & \omega_{|\Omega|-j+1}\notin F_i .
	\end{cases}
	\]
\end{theorem}

\begin{proof}
	We show the disjunctive case; the conjunctive case is obtained by replacing $\boldsymbol{C}^{\cup,\alpha}$ with $\boldsymbol{C}^{\cap,\alpha}$ and reversing the membership condition.  For a singleton $\{\omega\}$, only the bit corresponding to $\omega$ is changed by the elementary $\alpha$-junction operator; all other element bits are unchanged.  Hence the singleton operator is a tensor product in which one factor is $\boldsymbol{C}^{\cup,\alpha}$ and all remaining factors are $\boldsymbol{I}_2$.
	
	For a general focal set $F_i$, the singleton operators commute because they act on different element coordinates.  Their product is therefore
	\[
	\prod_{\omega\in F_i}\boldsymbol{K}^{\cup,\alpha}_{\{\omega\}}
	=\bigotimes_{j=1}^{|\Omega|}
	\begin{cases}
		\boldsymbol{C}^{\cup,\alpha}, & \omega_{|\Omega|-j+1}\in F_i,\\
		\boldsymbol{I}_2, & \omega_{|\Omega|-j+1}\notin F_i .
	\end{cases}
	\]
	This is the stated formula for $\boldsymbol{K}^{\cup,\alpha}_{F_i}$, and the same tensor-product argument gives the conjunctive formula.
\end{proof}

Theorem~\ref{theorem_alpha} gives a circuit decomposition of the $\alpha$-junction. Each two-dimensional factor acts on one element qubit, and their tensor product acts on the full register. Local controlled rotations followed by focal-set extraction implement the same operation without explicitly constructing the full matrices.

\begin{definition}\label{alpha_j_q_d}
	The conjunctive and disjunctive $\alpha$-junction rules are
		\begin{equation}\label{alpha_j_q_e1}
			\sum_{F_i\subseteq\Omega} m_1(F_i)\,\boldsymbol{K}^{\cap,\alpha}_{F_i}\,\boldsymbol{m}_2
			=\sum_{F_i\subseteq\Omega} m_1(F_i)\,\boldsymbol{m}^{\cap,\alpha}_{2,F_i},
		\end{equation}
		\begin{equation}
			\sum_{F_i\subseteq\Omega} m_1(F_i)\,\boldsymbol{K}^{\cup,\alpha}_{F_i}\,\boldsymbol{m}_2
			=\sum_{F_i\subseteq\Omega} m_1(F_i)\,\boldsymbol{m}^{\cup,\alpha}_{2,F_i}.
		\end{equation}
	Suppose $\ket{m_2}$ is encoded on $q_0,\dots,q_{n-1}$, with ancillas $q_{0_a},\dots,q_{(n-1)_a}$.
	
	\medskip
	\noindent\textbf{Conjunctive case.} The evolution is
	\[
	\begin{aligned}
		\ket{\psi_0} = \ket{m_2}\ket{0}^n, \ket{\psi_1} =
		X^{\otimes n}\ket{\psi_0}, ket{\psi_2} =
		\Bigl(\prod_{i=0}^{n-1} C^{\{q_i\}}R_Y(2\arccos\sqrt{\alpha})(q_{i_a})\Bigr)\ket{\psi_1}, \ket{\psi_3} =
		X^{\otimes n}\ket{\psi_2}
		\;\longrightarrow\;
		\ket{m_{2}^{\cap,\alpha}} .
	\end{aligned}
	\]
	\noindent\textbf{Disjunctive case.} The evolution is
	\[
	\begin{aligned}
		\ket{\psi_0} = \ket{m_2}\ket{0}^n, \ket{\psi_1} =
		\Bigl(\prod_{i=0}^{n-1} C^{\{q_i\}}R_Y(2\arccos\sqrt{\alpha})(q_{i_a})\Bigr)\ket{\psi_0}, \ket{\psi_2} =
		X^{\otimes n}_{q_{0_a},\dots,q_{n-1_a}}\ket{\psi_1}
		\;\longrightarrow\;
		\ket{m_{2}^{\cup,\alpha}} .
	\end{aligned}
	\]
	
	\medskip
	For a focal set $F_i$, introduce $q_{n_a},\dots,q_{(2n-1)_a}$ and apply
	\[
	\begin{cases}
		C^{\{q_{j-1_a}\}}X(q_{2j-1_a}), & \omega_j\notin F_i,\\
		C^{\{q_{j-1}\}}X(q_{2j-1_a}),   & \omega_j\in F_i,
	\end{cases}
	\]
	so that measuring $q_{n_a},\dots,q_{(2n-1)_a}$ yields $\boldsymbol{m}^{\cap,\alpha}_{2,F_i}$ or $\boldsymbol{m}^{\cup,\alpha}_{2,F_i}$.
\end{definition}

\begin{example}\label{ee3}
	Consider $\alpha=0.3$ and
	\[
		m=\{0.02,0.1,0.1,0.25,0.06,0.27,0.02,0.18\}.
	\]
	The state $\ket{m_{2}^{\cap,\alpha}}$ is prepared according to Definition~\ref{alpha_j_q_d}; the corresponding circuit is shown in Figure~\ref{alpha_junction_f}.
	\begin{figure}[H]
		\centering
		\includegraphics[width=0.85\textwidth]{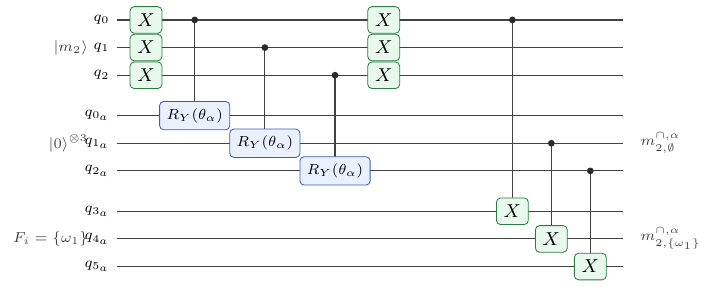}
		\caption{Complete partial implementation of $\ket{m^{\cap,0.3}}$ in Example~\ref{ee3}.  The same circuit prepares the conjunctive $\alpha$-state and extracts the focal-set registers associated with $m^{\cap,0.3}_{\emptyset}$ and $m^{\cap,0.3}_{\{\omega_1\}}$.}
		\label{alpha_junction_f}
	\end{figure}
	With 8096 shots, the measured probability vectors of the first and second three-qubit output registers are

		$$\begin{aligned}
			&[0.3657, 0.0461, 0.2205, 0.0345, 0.2297, 0.0493, 0.0473, 0.0069]^T,\\
			&[0.0678, 0.3367, 0.0462, 0.21, 0.0726, 0.2054, 0.0121, 0.0492]^T,
		\end{aligned}$$
respectively.  Eq.~(\ref{alpha_j_q_e1}) gives
	
		$$\begin{aligned}
			&m^{\cap,0.3}_{\emptyset}=\{0.3659, 0.0489, 0.2239, 0.0323, 0.2183, 0.0519, 0.0519, 0.0069\},\\
			&m^{\cap,0.3}_{\{\omega_1\}}=\{0.0698, 0.345, 0.0462, 0.21, 0.0742, 0.196, 0.0098, 0.049\}.
		\end{aligned}$$

\end{example}

\begin{theorem}\label{th4}
	For a fixed $\alpha$, the coefficient $m_{2,F_i}^{\cdot,\alpha}$ equals the squared amplitude of $\ket{m_{2,F_i}^{\cdot,\alpha}}$ in Definition~\ref{alpha_j_q_d}.
\end{theorem}

\begin{proof}
	For a one-element frame $\Omega=\{\omega\}$, the conjunctive circuit in Definition~\ref{alpha_j_q_d} applies an $X$ gate, a controlled $R_Y(\theta)$ with $\cos^2(\theta/2)=\alpha$, and a final $X$ gate.  If the queried focal set excludes $\omega$, the ancilla probabilities are
	\[
	\Pr[a=0]=\alpha m_2(\emptyset)+m_2(\{\omega\}),\qquad
	\Pr[a=1]=(1-\alpha)m_2(\emptyset),
	\]
	which are exactly the entries of $\boldsymbol{C}^{\cap,\alpha}\boldsymbol{m}_2$.  If the queried focal set contains $\omega$, the extraction gate copies the unchanged membership bit and gives the identity factor in Theorem~\ref{theorem_alpha}.  Since the multi-element case is the tensor product of these one-qubit factors, the squared amplitudes of $\ket{m_{2,F_i}^{\cap,\alpha}}$ equal the coefficients of $\boldsymbol{m}_{2,F_i}^{\cap,\alpha}$.  The disjunctive case follows from the same argument using $\boldsymbol{C}^{\cup,\alpha}$.
\end{proof}

This construction prepares a shared state $\ket{m_2^{\cdot,\alpha}}$ and extracts a requested intermediate vector $\boldsymbol{m}_{2,F_i}^{\cdot,\alpha}$ with controlled gates. The final weighted sum over $m_1(F_i)$ may be evaluated classically, making this a hybrid implementation of the matrix-calculus form.

A second implementation realizes the entire operation on a circuit with fewer gates. The conjunctive $\alpha$-junction interpolates between CECR and CCR, and the disjunctive form interpolates between DECR and DCR. Figure~\ref{tbm_qc_f1} retains the general layered BACR realization of CECR. For this equivalence predicate, Theorem~\ref{th2+1} gives the same output with two CNOT gates and one $X$ gate per element. Controlled $R_Y$ rotations then interpolate between the boundary circuits without constructing the matrices $\boldsymbol{K}^{\cdot,\alpha}_{F_i}$.

\begin{definition}\label{alpha_junction_q_d_2}
	Consider two mass functions $m_1$ and $m_2$, whose MFQSs are 
	$\ket{m_1}$ and $\ket{m_2}$ encoded on qubit registers 
	$q_{j_1},q_{j_2}$ with $j=0,\dots,n-1$.  
	The $\alpha$-junctions can be realized as follows.
	
	\noindent\textbf{Conjunctive case.}  
	Introduce $2n$ ancillas $q_{j_{a1}},q_{j_{a2}}$.  
	The state evolves as
	\[
	\begin{aligned}
		&\ket{\psi_0}=\ket{m_1}\ket{m_2}\ket{0}^{2n},\ket{\psi_1}=X^{\otimes 2n}_{\{q_{j_1},q_{j_2}\}}\ket{\psi_0},\\
		&\ket{\psi_2}=
		\Bigl(\prod_{j=0}^{n-1} C^{\{q_{j_1},q_{j_2}\}}
		X(q_{j_{a1}})\Bigr)\ket{\psi_1},\ket{\psi_3} = 
		\Bigl(\prod_{j=0}^{n-1} C^{\{q_{j_1}\}}X(q_{j_{a2}})\,
		C^{\{q_{j_2}\}}X(q_{j_{a2}})\Bigr)\ket{\psi_2}, \\[0.6em]
		&\ket{\psi_4} =
		\Bigl(\prod_{j=0}^{n-1} C^{\{q_{j_{a1}}\}}
		R_Y(2\arccos\sqrt{\alpha})(q_{j_{a2}})\Bigr)\ket{\psi_3},\ket{\psi_5}=X^{\otimes n}_{\{q_{j_{a2}}\}}\ket{\psi_4}
		\;\longrightarrow\;
		\ket{m_1 \circledsmall{$\cap$}^{\alpha} m_2}.
	\end{aligned}
	\]
	
	\noindent\textbf{Disjunctive case.}  
	Introduce again $2n$ ancillas $q_{j_{a1}},q_{j_{a2}}$.  
	The evolution is
	\[
		\begin{aligned}
			&\ket{\psi_0} = \ket{m_1}\ket{m_2}\ket{0}^{2n}, \ket{\psi_1} = 
			\Bigl(\prod_{j=0}^{n-1} C^{\{q_{j_1},q_{j_2}\}}X(q_{j_{a1}})\Bigr)\ket{\psi_0}, \ket{\psi_2} = 
			\Bigl(\prod_{j=0}^{n-1} C^{\{q_{j_1}\}}X(q_{j_{a2}})\,
			C^{\{q_{j_2}\}}X(q_{j_{a2}})\Bigr)\ket{\psi_1}, \\[0.6em]
			&\ket{\psi_3} =
			\Bigl(\prod_{j=0}^{n-1} C^{\{q_{j_{a1}}\}}R_Y(2\arccos\sqrt{\alpha})(q_{j_{a2}})\Bigr)\ket{\psi_2}
			\rightarrow
			\ket{m_1 \circledsmall{$\cup$}^{\alpha} m_2}.
		\end{aligned}
	\]
	
	Figure~\ref{alpha_junction_q_d_2_f} shows the complete circuit.
	\begin{figure}[htbp!]
		\centering
		\includegraphics[width=0.7\textwidth]{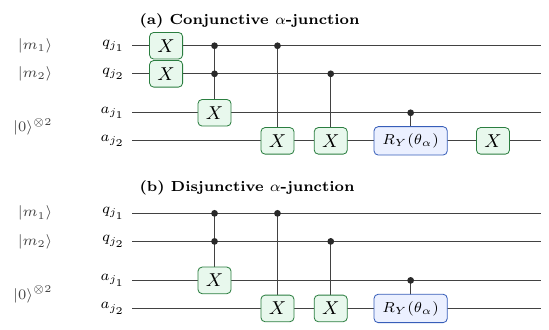}
		\caption{Full circuit implementation of the $\alpha$-junctions. In each panel, the input MFQS registers $\ket{m_1}$ and $\ket{m_2}$ precede the two ancillas initialized in $\ket{0}^{\otimes 2}$.}
		\label{alpha_junction_q_d_2_f}
	\end{figure}
\end{definition}

The two implementations serve different purposes. The hybrid form follows the original matrix definition: it applies belief revisions to one mass function and averages them with weights from the other. This form makes the revision semantics explicit. The full circuit interpolates between the two nonparameterized boundary rules and uses fewer gates. Table \ref{credal_level_com_t} compares their resource counts.

\subsection{Comparative resource analysis}

The constructions above give exact circuit semantics for credal-level operations. We compare classical arithmetic counts with logical-gate counts; these quantities do not represent wall-clock time. An end-to-end implementation must account for
\[
	C_{\mathrm{total}}
	=C_{\mathrm{prep}}+C_{\mathrm{op}}+C_{\mathrm{read}}
	+C_{\mathrm{compile}},
\]
Here $C_{\mathrm{prep}}$ denotes MFQS preparation, $C_{\mathrm{op}}$ the operation analyzed below, $C_{\mathrm{read}}$ sampling of the requested output, and $C_{\mathrm{compile}}$ gate decomposition, routing, and error-control overhead. For an arbitrary mass function, $C_{\mathrm{prep}}$ can scale as $O(2^n)$ (Section~\ref{method1}). Reconstructing all $2^n$ output masses likewise requires exponentially many output coordinates or samples. The results below characterize operation subroutines on prepared MFQSs and do not imply an unconditional end-to-end speedup.

\begin{remark}
	The quantum counts below are logical-gate inventories. Toffoli, CNOT, and controlled-$R_Y$ gates are reported separately because they have different decomposition and fault-tolerant costs. Circuit depth additionally depends on qubit connectivity, parallel scheduling, and the available work qubits.
\end{remark}

\begin{theorem}\label{th5}
	Let $m_1,\dots,m_k$ be $k$ mass functions defined on an $n$-element FoD, with MFQSs $\ket{m_1},\dots,\ket{m_k}$.  
	If the BACR corresponds to $\mathfrak{B}=F_{i_1}\cap\cdots\cap F_{i_k}$, then a clean-workspace implementation of the multi-source CCR uses $(2k-3)n$ Toffoli gates and $(k-2)n$ reusable work qubits. If intermediate conjunctions may remain as garbage, the forward computation uses $(k-1)n$ Toffoli gates.
\end{theorem}

\begin{proof}
	For each element $\omega_j$, the $k$ membership bits control one output bit. Using $k-2$ clean work qubits, $k-2$ Toffoli gates first compute the intermediate conjunctions and one additional Toffoli writes the final conjunction to the output. Restoring the work qubits requires another $k-2$ Toffoli gates, giving $2k-3$ gates per element. Omitting this uncomputation leaves the intermediate conjunctions in the work register and uses only $k-1$ gates. Applying the construction independently to all $n$ elements gives the stated counts \cite{barenco1995elementary}.
\end{proof}

\begin{theorem}\label{th6}
	Assume that the input contour coordinates of poss-transferable mass functions, or the input commonality coordinates of general mass functions, are already available. The pointwise product stage of the classical CCR requires $(k-1)n$ multiplications in the former case and $(k-1)2^n$ multiplications in the latter.
\end{theorem}

\begin{proof}\label{th6_proof}
	For poss-transferable mass functions, CCR is computed through element-wise contour multiplication:
	\[
	pl_{1\circledtiny{$\cap$}\cdots\circledtiny{$\cap$}k}(\omega_j)
	= \prod_{\ell=1}^{k}pl_\ell(\omega_j),\qquad j=1,\dots,n .
	\]
	This requires $(k-1)n$ multiplications.  For a general mass function, CCR is computed over all commonality coordinates:
	\[
	q_{1\circledtiny{$\cap$}\cdots\circledtiny{$\cap$}k}(F)
	= \prod_{\ell=1}^{k}q_\ell(F),\qquad F\subseteq\Omega .
	\]
	There are $2^n$ focal-set coordinates, so the product stage costs $(k-1)2^n$ multiplications. These counts exclude transformations between mass, contour, and commonality representations and exclude the cost of materializing the final mass vector.
\end{proof}

Theorems~\ref{th5} and \ref{th6} have different outputs. The classical subroutine explicitly updates every requested coordinate; the quantum subroutine leaves its result as an MFQS. Both have linear operation counts for poss-transferable mass functions. For general mass functions, the prepared-state circuit evaluates focal-set intersection on $n$ element positions. Recovering the complete output mass function still requires exponentially many coordinates. The circuit reduction is therefore relevant when MFQS preparation is efficient and the task uses a limited number of belief queries or samples.

\begin{theorem}\label{th7}
	Consider a mass function $m$ under an $n$-element FoD, with its MFQS represented as $\ket{m}$ in an $n$-qubit system. Implementing $\ket{{m}^{\cdot,\alpha}}$ requires $n$ controlled-$R_Y$ gates. Extracting all states $\ket{{m}^{\cdot,\alpha}_{F_0}},\ldots,\ket{{m}^{\cdot,\alpha}_{F_{2^n-1}}}$ requires $n2^n$ CNOT gates; retaining all outputs simultaneously also requires $n2^n$ output qubits.
\end{theorem}

\begin{proof}
	The construction of $\ket{m^{\cdot,\alpha}}$ in Definition~\ref{alpha_j_q_d} applies one controlled-$R_Y$ rotation to each element qubit. Extracting the component associated with one focal set $F_i$ then requires one CNOT and one output qubit per element coordinate. Repeating the extraction for all $2^n$ focal sets gives the stated gate and output-register counts. Sequential extraction can reuse the output register, but then requires repeated circuit executions.
\end{proof}

\begin{theorem}\label{th7+1}
	Consider two mass functions, $m_1$ and $m_2$, under an $n$-element frame of discernment, with their MFQSs represented as $\ket{m_1}$ and $\ket{m_2}$. The $\alpha$-junction circuit in Definition~\ref{alpha_junction_q_d_2} uses $n$ Toffoli-type gates, $2n$ CNOT gates, and $n$ controlled-$R_Y$ gates.
\end{theorem}

\begin{proof}
	For each element coordinate, Definition~\ref{alpha_junction_q_d_2} uses one Toffoli-type gate for the CCR/DCR branch, two CNOT gates for the CECR/DECR branch, and one controlled-$R_Y$ gate for interpolation. Summing each gate type over the $n$ coordinates gives the stated inventory.
\end{proof}

\begin{theorem}\label{th8}
	In a direct coordinate-wise classical construction, generating all $\boldsymbol{m}^{\cdot,\alpha}_{F_i}$ requires $n2^n$ scalar multiplications for a poss-transferable mass function and $2^{2n}$ scalar multiplications for a general mass function.
\end{theorem}

\begin{proof}
	For a poss-transferable mass function, each $\boldsymbol{m}^{\cdot,\alpha}_{F_i}$ can be obtained by modifying $n$ contour coordinates.  Repeating this for all $2^n$ focal sets gives $n2^n$ multiplications.  For a general mass function, the transformation acts on full vectors indexed by $2^n$ focal sets.  Constructing all $\boldsymbol{m}^{\cdot,\alpha}_{F_i}$ therefore requires accounting for $2^n$ input coordinates for each of $2^n$ queried focal sets, yielding $2^{2n}$ scalar multiplications in the direct classical implementation.
\end{proof}

\begin{table}[H]
	\centering
	\caption{Subroutine resource counts at the credal level. Classical and quantum entries report different primitive types and are not wall-clock-time comparisons.}
	\label{credal_level_com_t}
	\resizebox{0.98\linewidth}{!}{
		\begin{tabular}{lll}
			\Xhline{1.5pt}
			Subroutine & \makecell[l]{Classical coordinate\\operations} & \makecell[l]{MFQS circuit gates for prepared inputs} \\
			\hline
			\makecell[l]{Poss-transferable mass\\function (CCR)} & $(k-1)n$ multiplications & \makecell[l]{$(2k-3)n$ Toffoli gates with clean work qubits} \\
			General mass function (CCR) & $(k-1)2^n$ multiplications & $(2k-3)n$ Toffoli gates with clean work qubits \\
			\makecell[l]{Poss-transferable mass function\\(partial $\alpha$-junction)} & $n2^n$ multiplications & \makecell[l]{$n$ controlled-$R_Y$ and $n2^n$ CNOT gates} \\
			General mass function (partial $\alpha$-junction) & $2^{2n}$ multiplications & $n$ controlled-$R_Y$ and $n2^n$ CNOT gates \\
			\makecell[l]{Poss-transferable mass function\\(entire $\alpha$-junction)} & $n2^n$ multiplications & \makecell[l]{$n$ Toffoli-type, $2n$ CNOT and $n$ controlled-$R_Y$ gates} \\
			\makecell[l]{General mass function\\(entire $\alpha$-junction)} & $2^{2n}$ multiplications & \makecell[l]{$n$ Toffoli-type, $2n$ CNOT and $n$ controlled-$R_Y$ gates} \\
			\Xhline{1.5pt}
		\end{tabular}}
\end{table}

Table~\ref{credal_level_com_t} summarizes the subroutine counts. Prepared general MFQSs reduce focal-set-indexed updates at the operation stage. End-to-end savings further require efficient state preparation, query-level output, and manageable compilation costs. The direct Boolean semantics on focal-set membership bits also distinguishes this construction from generic linear-system and variational implementations \cite{low2014quantum,zhou2023bf,luo2024variational}.

Table~\ref{credal_level_com_t} also shows how MFQS encoding changes the operation load. Repeated gate modules on element qubits replace explicit iteration over focal-set coordinates. For prepared general MFQSs, CCR and the entire $\alpha$-junction have $O(n)$ logical-gate inventories; direct classical coordinate constructions use $2^n$ or $2^{2n}$ focal-set-indexed operations. This reduction is useful for tasks that consume samples or a small set of belief queries. Materializing the complete mass vector reintroduces exponential readout cost.

\section{Operations on quantum circuits product space}
\label{method3}

The preceding sections operate within one FoD. Product-space operations project, replicate, or conditionally embed a focal set on $\Omega\times\Theta$. Under MFQS encoding, one qubit represents each composite element $(\omega_i,\theta_j)$. Controlled-$X$ gates therefore implement marginalization, vacuous extension, and ballooning extension as Boolean transformations of membership bits.

\subsection{Marginalization on quantum circuits}

\begin{definition}\label{marginal_q_d}
	Consider a mass function on the product frame $\Omega\times\Theta$, denoted $m^{\Omega\times\Theta}$, whose MFQS $\ket{m^{\Omega\times\Theta}}$ is implemented in a $|\Omega|\times|\Theta|$-qubit register 
	$q_{(0,0)},q_{(0,1)},\dots,q_{(|\Omega|-1,|\Theta|-1)}$,  
	where $q_{(i,j)}$ encodes the element $(\omega_{i+1},\theta_{j+1})$.  
	The marginalization of $m^{\Omega\times\Theta}$ on $\Omega$ can be realized as follows.
	\[
	\begin{aligned}
		\ket{\psi_0}=\ket{m^{\Omega\times\Theta}}\ket{0}^{|\Omega|},\ket{\psi_1}=
		\Bigl(\prod_{i=0}^{|\Omega|-1}
		C^{\{\overline{q_{(i,0)}},\dots,\overline{q_{(i,|\Theta|-1)}}\}}
		X(q_{i_a})\Bigr)\ket{\psi_0},\ket{m^{\Omega\times\Theta\downarrow\Omega}}=
		X^{\otimes|\Omega|}_{\{q_{0_a},\dots,q_{(|\Omega|-1)_a}\}}\ket{\psi_1}.
	\end{aligned}
	\]
	Here an overlined control is activated by state $\ket{0}$. The two steps compute
	$q_{i_a}=\bigvee_{j=0}^{|\Theta|-1}q_{(i,j)}$, so the ancilla register stores whether each $\omega_{i+1}$ appears in the product-space focal set.
\end{definition}

\begin{example}\label{e3}
	Consider the following mass function on $\Omega\times\Theta$:
	\[
	\begin{aligned}
		&m^{\Omega\times\Theta}(\{(\omega_1,\theta_1)\})=0.1,m^{\Omega\times\Theta}(\{(\omega_1,\theta_1),(\omega_1,\theta_2)\})=0.2,\\
		&m^{\Omega\times\Theta}(\{(\omega_1,\theta_2),(\omega_2,\theta_2)\})=0.4,m^{\Omega\times\Theta}(\Omega\times\Theta)=0.3 .
	\end{aligned}
	\]
	By Eq.~\eqref{margin_e}, its marginalizations on $\Omega$ and $\Theta$ are
	\[
	\begin{aligned}
		m^{\Omega\times\Theta\downarrow\Omega}(\{\omega_1\})=0.3,\quad
		m^{\Omega\times\Theta\downarrow\Omega}(\Omega)=0.7,m^{\Omega\times\Theta\downarrow\Theta}(\{\theta_1\})=0.1,\quad
		m^{\Omega\times\Theta\downarrow\Theta}(\{\theta_2\})=0.4,\quad
		m^{\Omega\times\Theta\downarrow\Theta}(\Theta)=0.5.
	\end{aligned}
	\]
	Figure~\ref{marginal_qc_f} shows the corresponding circuits. With output registers ordered as $(\omega_1,\omega_2)$ and $(\theta_1,\theta_2)$, the nonzero measured probabilities are
	\[
	\begin{aligned}
		\Omega\text{-register:} \Pr(\ket{10})=0.3,\Pr(\ket{11})=0.7,\quad\Theta\text{-register:}\Pr(\ket{10})=0.1, \Pr(\ket{01})=0.4,\Pr(\ket{11})=0.5 .
	\end{aligned}
	\]
	\begin{figure}[htbp!]
		\centering
		\includegraphics[width=0.92\textwidth]{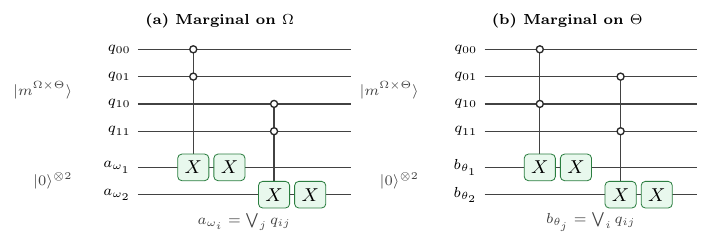}
		\caption{Circuit implementation of marginalization in Example~\ref{e3}. The four-qubit input register $\ket{m^{\Omega\times\Theta}}$ is ordered before the two output ancillas initialized in $\ket{0}^{\otimes 2}$.}
		\label{marginal_qc_f}
	\end{figure}
\end{example}

\begin{theorem}
	The squared amplitudes of the output state $\ket{m^{\Omega\times\Theta\downarrow\Omega}}$ in Definition~\ref{marginal_q_d} equal the outcome of marginalization in Eq.~\eqref{margin_e}.
\end{theorem}

\begin{proof}
	Write the MFQS as
	\[
	\ket{m^{\Omega\times\Theta}}=\sum_{H\subseteq\Omega\times\Theta}
	\sqrt{m^{\Omega\times\Theta}(H)}\ket{x_H},
	\]
	where $x_H(i,j)=1$ iff $(\omega_{i+1},\theta_{j+1})\in H$. For each $\omega_{i+1}$, the open-controlled $X$ in Definition~\ref{marginal_q_d} first detects the case
	$x_H(i,0)=\cdots=x_H(i,|\Theta|-1)=0$. The final $X$ changes this all-zero test into
	\[
	y_i=\bigvee_{j=0}^{|\Theta|-1}x_H(i,j).
	\]
	Thus the output ancilla string is exactly the membership vector of the projection
	$H^{\downarrow\Omega}=\{\omega_{i+1}:\exists\,\theta_{j+1},(\omega_{i+1},\theta_{j+1})\in H\}$.
	Measuring an output focal set $F\subseteq\Omega$ therefore gives
	\[
	\Pr(F)=\sum_{H:\,H^{\downarrow\Omega}=F}m^{\Omega\times\Theta}(H),
	\]
	which is precisely the marginal mass $m^{\Omega\times\Theta\downarrow\Omega}(F)$ in Eq.~\eqref{margin_e}.
\end{proof}

\subsection{Vacuous extension on quantum circuits}

\begin{definition}\label{vacuous_q_d}
	Consider a mass function $m^{\Omega}$ under the FoD $\Omega$, with MFQS $\ket{m^{\Omega}}$ implemented in an $|\Omega|$-qubit system 
	$q_0,\dots,q_{|\Omega|-1}$,  
	where $q_i$ corresponds to element $\omega_{i+1}$.  
	The vacuous extension of $m^{\Omega}$ to $\Omega\times\Theta$ can be realized as follows.
	\[
	\begin{aligned}
		&\ket{\psi_0}=\ket{m^{\Omega}}\ket{0}^{|\Omega|\cdot|\Theta|}, \ket{m^{\Omega\uparrow\Omega\times\Theta}}=
		\Bigl(\prod_{i=0}^{|\Omega|-1}\prod_{j=0}^{|\Theta|-1} 
		C^{\{q_i\}}X(q_{(i,j)_a})\Bigr)\ket{\psi_0}, 
	\end{aligned}
	\]
	where $q_{(i,j)_a}$ denotes the state of the composite element $(\omega_{i+1},\theta_{j+1})\in\Omega\times\Theta$.
\end{definition}

\begin{example}\label{e4}
	Consider a mass function under the FoD $\Omega$,
	\[
	m^{\Omega}(\{\omega_1\})=0.1,\quad
	m^{\Omega}(\{\omega_2\})=0.4,\quad
	m^{\Omega}(\Omega)=0.5.
	\]
	By Eq.~\eqref{vacuous_e}, its vacuous extension on $\Omega\times\Theta$ assigns the same masses to $\{\omega_1\}\times\Theta$, $\{\omega_2\}\times\Theta$, and $\Omega\times\Theta$, respectively. Figure~\ref{vacuous_extension_q_f} shows the circuit. For $|\Omega|=|\Theta|=2$ and output order $((\omega_1,\theta_1),(\omega_1,\theta_2),(\omega_2,\theta_1),(\omega_2,\theta_2))$, the nonzero measured probabilities are
	\[
	\Pr(\ket{1100})=0.1,\quad
	\Pr(\ket{0011})=0.4,\quad
	\Pr(\ket{1111})=0.5.
	\]
	\begin{figure}[htbp!]
		\centering
		\includegraphics[width=0.5\textwidth]{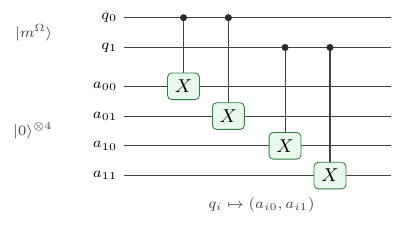}
		\caption{Circuit implementation of vacuous extension in Example~\ref{e4}. The input register $\ket{m^{\Omega}}$ is ordered before the four product-space output qubits initialized in $\ket{0}^{\otimes 4}$.}
		\label{vacuous_extension_q_f}
	\end{figure}
\end{example}

\begin{theorem}
	The squared amplitudes of the output state $\ket{m^{\Omega\uparrow\Omega\times\Theta}}$ in Definition~\ref{vacuous_q_d} equal the outcome of vacuous extension in Eq.~(\ref{vacuous_e}).
\end{theorem}

\begin{proof}
	Write the input MFQS as
	\[
	\ket{m^{\Omega}}=\sum_{F\subseteq\Omega}\sqrt{m^\Omega(F)}\ket{x_F}.
	\]
	The circuit in Definition~\ref{vacuous_q_d} copies each membership bit $x_F(i)$ to all ancillas $q_{(i,0)_a},\dots,q_{(i,|\Theta|-1)_a}$ by controlled-$X$ gates. Therefore
	\[
	\ket{x_F}\ket{0}^{|\Omega||\Theta|}
	\longmapsto
	\ket{x_F}\ket{x_{F\times\Theta}}.
	\]
	No amplitude is changed; only the support label is re-encoded from $F$ to $F\times\Theta$. Hence the probability of measuring the product-space focal set $F\times\Theta$ is $m^\Omega(F)$, and all other focal sets receive zero mass. This is exactly Eq.~\eqref{vacuous_e}.
\end{proof}

\subsection{Ballooning extension on quantum circuits}

\begin{definition}\label{ballooing_q_d}
	Consider a mass function with a fixed subset $F_i\subseteq\Theta$ under an FoD $\Omega$, denoted $m^{\Omega}[F_i]$.  
	Its MFQS $\ket{m^{\Omega}[F_i]}$ is implemented in an $|\Omega|$-qubit register 
	$q_0,\dots,q_{|\Omega|-1}$, where $q_k$ corresponds to element $\omega_{k+1}$.  
	The ballooning extension of $m^{\Omega}[F_i]$ to $\Omega\times\Theta$ can be realized as follows.
	\[
	\begin{aligned}
		\ket{\psi_0} = \ket{m^{\Omega}[F_i]}\ket{0}^{|\Omega|\cdot|\Theta|}, \ket{\psi_1} =
		\Bigl(\prod_{\theta_j\in F_i}\prod_{k=0}^{|\Omega|-1} 
		C^{\{q_k\}}X(q_{(k,j)_a})\Bigr)\ket{\psi_0}, \ket{m^{\Omega}[F_i]^{\Uparrow\Omega\times\Theta}} =
		\Bigl(\prod_{\theta_j\notin F_i} 
		X^{\otimes|\Omega|}_{\{q_{(0,j)_a},\dots,q_{(|\Omega|-1,j)_a}\}}\Bigr)\ket{\psi_1},
	\end{aligned}
	\]
	where $q_{(k,j)_a}$ encodes the composite element $(\omega_{k+1},\theta_{j+1})\in\Omega\times\Theta$.
\end{definition}

\begin{example}\label{e5}
	Consider a mass function with given $\theta_2$ under the FoD $\Omega$,
	\[
	m^{\Omega}[\{\theta_2\}](\{\omega_1\})=0.1,\quad
	m^{\Omega}[\{\theta_2\}](\{\omega_2\})=0.4,\quad
	m^{\Omega}[\{\theta_2\}](\Omega)=0.5.
	\]
	According to Eq.~\eqref{ballooning_e}, its ballooning extension on $\Omega\times\Theta$ is
	\[
	\begin{aligned}
		&m^{\Omega}[\{\theta_2\}]^{\Uparrow\Omega\times\Theta}((\omega_1,\theta_2),(\omega_2,\theta_1),(\omega_2,\theta_2))=0.1,m^{\Omega}[\{\theta_2\}]^{\Uparrow\Omega\times\Theta}((\omega_1,\theta_1),(\omega_1,\theta_2),(\omega_2,\theta_2))=0.4,m^{\Omega}[\{\theta_2\}]^{\Uparrow\Omega\times\Theta}(\Omega\times\Theta)=0.5.
	\end{aligned}
	\]
	Figure~\ref{ballooning_extension_q_f} implements Definition~\ref{ballooing_q_d}. The nonzero measured probabilities of $q_{(0,0)_a}\cdots q_{(1,1)_a}$ are
	\[\begin{aligned}
		&\Pr(\ket{1110})=0.1,\quad \Pr(\ket{1011})=0.4,\quad \Pr(\ket{1111})=0.5.
	\end{aligned}\]
	\begin{figure}[htbp!]
		\centering
		\includegraphics[width=0.65\textwidth]{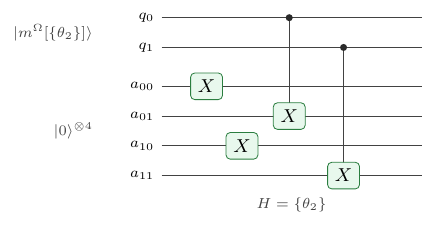}
		\caption{Circuit implementation of ballooning extension in Example~\ref{e5}. The input register $\ket{m^{\Omega}[\{\theta_2\}]}$ is ordered before the four product-space output qubits initialized in $\ket{0}^{\otimes 4}$.}
		\label{ballooning_extension_q_f}
	\end{figure}
\end{example}

\begin{theorem}
	The squared amplitudes of the output state $\ket{m^{\Omega}[\{\theta_2\}]^{\Uparrow\Omega\times\Theta}}$ in Definition~\ref{ballooing_q_d} equal the outcome of ballooning extension in Eq.~\eqref{ballooning_e}.
\end{theorem}

\begin{proof}
	Let the conditioning subset in $\Theta$ be denoted by $H_\Theta$. For an input focal set $G\subseteq\Omega$, the MFQS branch is $\ket{x_G}$. Definition~\ref{ballooing_q_d} sets each output bit by
	\[
	y_{k,j}=
	\begin{cases}
		x_G(k), & \theta_{j+1}\in H_\Theta,\\
		1, & \theta_{j+1}\notin H_\Theta .
	\end{cases}
	\]
	Therefore the output focal set represented by the ancilla register is $(G\times H_\Theta)\cup(\Omega\times\overline{H_\Theta})$. The circuit changes only the computational-basis label of each branch and preserves its amplitude. Hence the probability of this output focal set is $m^\Omega[H_\Theta](G)$, which is exactly the mass assigned by the ballooning extension in Eq.~\eqref{ballooning_e}.
\end{proof}

\subsection{Generalized Bayesian theorem on quantum circuits}

The product-space circuits support valuation-based and generalized Bayesian reasoning. A VBS combines evidence from different FoDs and marginalizes the result \cite{shenoy1990axioms}. The GBT embeds conditional evidence by ballooning extension and updates it by Dempster's conditioning \cite{smets1993belief}. MFQS implementations of these primitives keep both procedures in the focal-set representation.

\begin{figure}[H]
	\centering
	\includegraphics[width=0.96\textwidth]{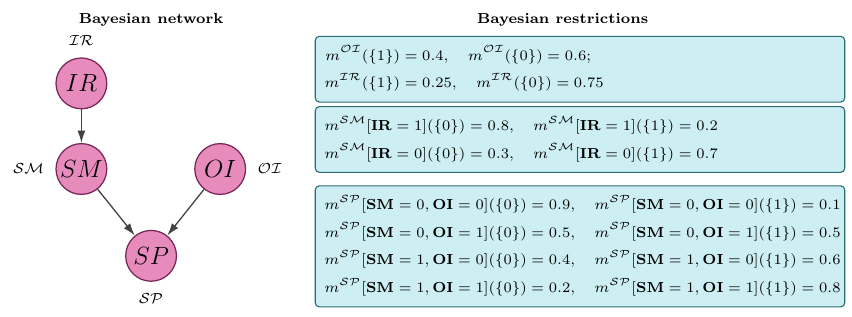}
	\caption{Four-node Bayesian network and its local Bayesian restrictions rendered in the same visual style as the valuation network in Figure~\ref{vbs_bayesian_f}.}
	\label{dag_fig}
\end{figure}

We illustrate the construction with the four-node Bayesian network from \cite{borujeni2021quantum}. Its binary variables \textbf{IR}, \textbf{OI}, \textbf{SM}, and \textbf{SP} have frames $\mathcal{IR}$, $\mathcal{OI}$, $\mathcal{SM}$, and $\mathcal{SP}$, respectively. Figure~\ref{dag_fig} gives the network and its local Bayesian mass functions.

Borujeni \textit{et al.} \cite{borujeni2021quantum} implemented this network in a probabilistic quantum framework based on Low's method \cite{low2014quantum}. The TBM circuit expresses the same inference through MFQS preparation, ballooning extension, combination, and marginalization. Low's construction assigns one qubit to each binary random variable. MFQS assigns qubits to frame elements and thereby also represents non-Bayesian mass functions with multi-element focal sets.

\begin{figure}[htbp!]
	\centering
	\includegraphics[width=0.6\textwidth]{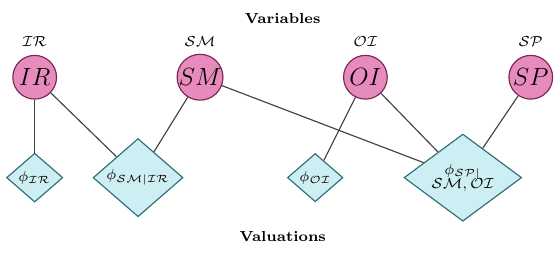}
	\caption{Valuation-network representation of the four-node Bayesian network. Circular nodes are variables, diamond nodes are valuations, and each edge links a valuation to a variable in its scope.}
	\label{vbs_bayesian_f}
\end{figure}

Figure~\ref{vbs_bayesian_f} expresses the Bayesian network as a valuation network. Each local conditional table is a valuation on its family scope. Valuation combination gives the joint distribution, and projection gives the requested marginal \cite{shenoy1992valuation,shenoy1990axioms}. The four-node network is therefore $\Phi=\phi_{\mathcal{IR}}\otimes\phi_{\mathcal{OI}}\otimes\phi_{\mathcal{SM}\mid\mathcal{IR}}\otimes\phi_{\mathcal{SP}\mid\mathcal{SM},\mathcal{OI}}$; a query on $\mathcal{SP}$ returns $\Phi^{\downarrow\mathcal{SP}}$.

\begin{figure}[htbp!]
	\centering
	\includegraphics[width=0.96\textwidth]{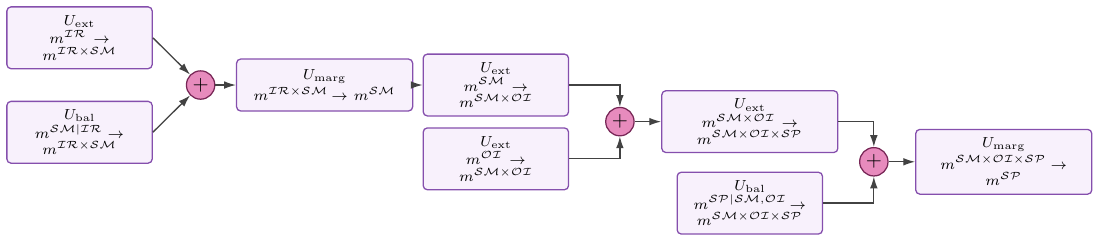}
	\caption{TBM-circuit workflow for the four-node GBT example, rendered with the same color and typography system as the VBS representation. Purple blocks are MFQS product-space primitives, and plus nodes denote Dempster combination.}
	\label{gbt_qc_f}
\end{figure}

Figure~\ref{gbt_qc_f} shows the corresponding TBM circuit. The blocks $U_{\mathrm{ext}}$, $U_{\mathrm{bal}}$, and $U_{\mathrm{marg}}$ implement vacuous extension, ballooning extension, and marginalization; plus nodes denote Dempster combination. The circuit extends the prior on $\mathcal{IR}$ and the conditional mass $m^{\mathcal{SM}\mid\mathcal{IR}}$ to $\mathcal{IR}\times\mathcal{SM}$, combines them, and marginalizes to obtain $m^{\mathcal{SM}}$. It then combines $m^{\mathcal{SM}}$ with $m^{\mathcal{OI}}$, embeds the conditional evidence for $\mathcal{SP}$, and marginalizes to $m^{\mathcal{SP}}$.

The construction assembles the network from reusable belief function primitives. Because each MFQS branch encodes a focal set, the workflow also accepts restrictions that assign mass to multi-element sets. A purely binary Bayesian network may require more element qubits than Low's one-qubit-per-variable construction, so we claim no general reduction in qubit count. The contribution is semantic: Bayesian inference becomes a special case of focal-set manipulation that preserves ignorance, imprecision, and conditional belief evidence.

\section{A multi-domain fusion example under gate-control noise}
\label{sec:noise_experiment}

\subsection{Motivation and scenario setting}

Quantum circuits provide a developing platform for multi-domain information fusion \cite{qu2024qmfnd,li2025qmlsc}. We evaluate the MFQS encoding in Eq.~(\ref{q_bpa}) and the conjunctive circuit in Definition~\ref{bacr_qc_d} in a noisy learning scenario. The experiment tests whether focal-set fusion preserves interpretable and reliable decision masses when gate-control noise perturbs the evidence QNNs.

We divide MNIST into two five-class tasks. Task A contains digits $\{0,1,2,3,4\}$, and Task B contains $\{5,6,7,8,9\}$. In each task, the five classes form the FoD $\Theta=\{c_1,\ldots,c_5\}$. The same image supplies two source domains. QNN-E1 receives the time-domain image. QNN-E2 receives a frequency-domain representation produced by a two-dimensional Fourier transform, spectrum centering, logarithmic magnitude extraction, and normalization. Both sources use the same compact convolutional amplitude encoder and six-layer strongly entangling evidence QNN. The normalized encoder output is loaded into a five-qubit register, and the QNN measurement distribution defines a mass function on $2^\Theta$ through Eq.~(\ref{q_bpa}).

Let $m_t(\cdot \mid x)$ and $m_f(\cdot \mid x)$ denote the time-domain and frequency-domain masses. Equation~(\ref{ccr_e}) combines them conjunctively and retains the conflict mass. BetP then gives the class decision. The fusion stage uses no $\alpha$ discounting or downstream QNN, so each decision remains traceable to focal-set intersection.

The comparison method is a CNOT-only All-to-All joint model. It uses the same evidence architectures and MFQS outputs as Eq.~(\ref{q_bpa}), with a fixed ten-qubit CNOT block in place of focal-set intersection. The fusion block has no trainable parameters. The two five-qubit evidence QNNs are trained jointly through its ten-qubit objective. After fusion, the last five qubits are marginalized out and BetP is applied to the first five-qubit marginal. This model couples the sources during training without explicitly imposing $C=A\cap B$.

\begin{figure}[htbp!]
	\centering
	\includegraphics[width=0.9\textwidth]{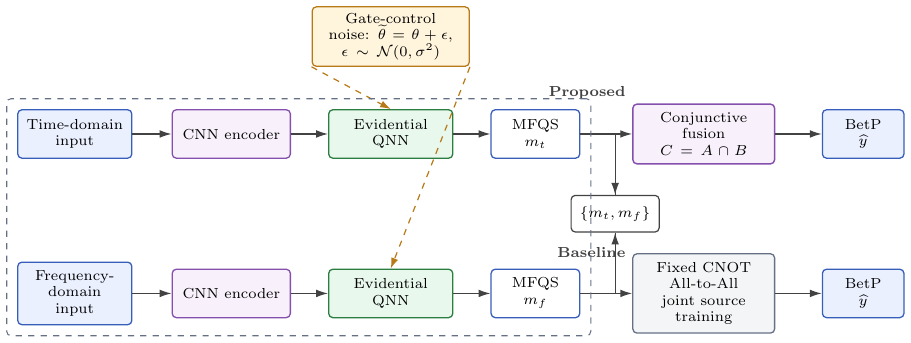}
	\caption{Experimental workflow for noisy evidential fusion. Angle noise is injected into the parameterized rotations of the two evidence QNNs before the MFQS masses are formed. The proposed path fuses the same noisy masses by focal-set intersection, whereas the comparison path uses a fixed CNOT All-to-All block.}
	\label{fig:noisy_evidential_fusion_workflow}
\end{figure}

Gate-control noise is simulated at inference time by perturbing every parameterized rotation angle in the evidence QNNs:
\[
	\tilde{\theta}=\theta+\epsilon,\qquad
	\epsilon\sim \mathcal{N}(0,\sigma^2).
\]
We scan $\sigma\in[0,0.34]$ and report the mean accuracy over five noise realizations at each nonzero level. The perturbation affects all parameterized rotations in the two evidence QNNs. Both fusion stages are fixed: focal-set intersection defines the proposed method, and CNOT gates define the All-to-All model. We do not perturb the CNOT gates in this simulation. Two-qubit hardware errors require a separate hardware-oriented study. Figure~\ref{fig:noisy_evidential_fusion_workflow} summarizes the protocol; both fusion methods receive the same noisy time-domain and frequency-domain masses.

\subsection{Experiments}

Table~\ref{tab:noise_clean_results} reports clean test accuracy. On Task A, conjunctive fusion reaches $99.64\%$, compared with $99.48\%$ for QNN-E1, $98.24\%$ for QNN-E2, and $99.20\%$ for the CNOT-only model. On Task B, it matches QNN-E1 at $98.92\%$ and slightly exceeds the CNOT-only model. QNN-E2 is weaker at $91.64\%$.

\begin{table}[htbp!]
	\centering
	\caption{Clean test accuracy of the single-source and fusion variants.}
	\label{tab:noise_clean_results}
	\resizebox{0.65\linewidth}{!}{
		\begin{tabular}{ccccc}
			\Xhline{1.5pt}
			Task & QNN-E1 & QNN-E2  & CNOT All-to-All& Conjunctive fusion \\
			\hline
			A: digits $0$--$4$ & 0.9948 & 0.9824  & 0.9920& \textbf{0.9964} \\
			B: digits $5$--$9$ & \textbf{0.9892} & 0.9164  & 0.9880& \textbf{0.9892} \\
			\Xhline{1.5pt}
	\end{tabular}}
\end{table}

\begin{figure}[htbp!]
	\centering
	\includegraphics[width=0.95\textwidth]{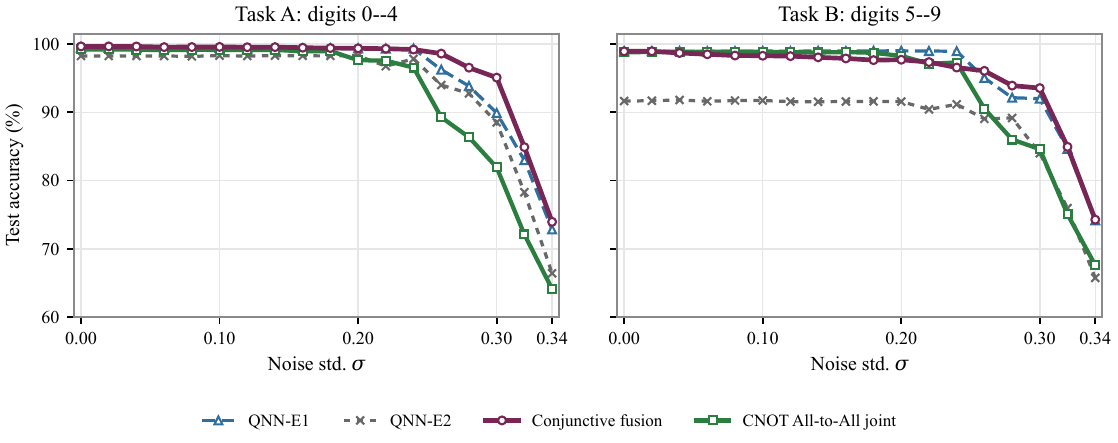}
	\caption{Noise robustness comparison among the two single-source QNNs, the proposed conjunctive fusion rule, and the CNOT-only All-to-All joint baseline.}
	\label{fig:noise_robustness_comparison}
\end{figure}

Figure~\ref{fig:noise_robustness_comparison} compares robustness across the four methods. On Task A at $\sigma=0.28$, conjunctive fusion reaches $96.53\%$, compared with $93.84\%$ for QNN-E1, $92.75\%$ for QNN-E2, and $86.35\%$ for the CNOT-only model. At $\sigma=0.34$, the corresponding accuracies are $73.96\%$, $72.84\%$, $66.42\%$, and $64.13\%$.

Task B gives a less uniform comparison because its frequency-domain source is weaker. The CNOT-only model is slightly more accurate at $\sigma=0.20$ and $\sigma=0.24$. At $\sigma=0.28$, conjunctive fusion reaches $93.90\%$, compared with $92.13\%$ for QNN-E1, $89.17\%$ for QNN-E2, and $86.01\%$ for the CNOT-only model. At $\sigma=0.34$, it reaches $74.28\%$, close to QNN-E1 at $74.18\%$ and above the other two methods.

The CNOT-only model is competitive in the clean and low-noise regimes, then degrades more rapidly as angle noise increases. The fixed fusion block receives no angle perturbation, so the difference arises from the learned source representations. Joint training must produce representations that remain useful after All-to-All CNOT mixing. Conjunctive fusion directly intersects the two source MFQSs.

\begin{figure}[H]
	\centering
	\includegraphics[width=0.95\textwidth]{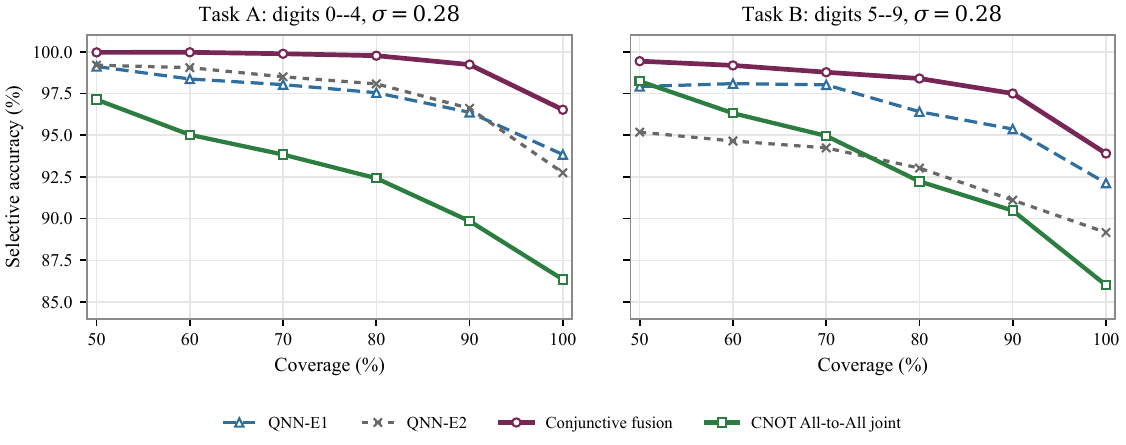}
	\caption{Selective classification at $\sigma=0.28$. Samples are ranked by the maximum BetP value, and lower-confidence samples are rejected. The proposed conjunctive fusion gives the strongest selective accuracy on both tasks and maintains a clearer margin over the CNOT-only All-to-All baseline as coverage decreases.}
	\label{fig:selective_rejection_sigma028}
\end{figure}

We evaluate selective classification at $\sigma=0.28$. Samples are ranked by maximum BetP, and a specified top-confidence fraction is accepted. This common rejection rule measures whether confidence identifies unreliable decisions. Table~\ref{tab:selective_rejection_results} reports accuracy at full, $80\%$, and $60\%$ coverage together with the AURC. Figure~\ref{fig:selective_rejection_sigma028} shows the corresponding curves.

\begin{table}[H]
	\centering
	\caption{Selective classification results at $\sigma=0.28$. Accuracy values are shown in percent; AURC is lower when the confidence ranking is better aligned with correctness.}
	\label{tab:selective_rejection_results}
	\resizebox{0.85\linewidth}{!}{
		\begin{tabular}{cccccc}
			\Xhline{1.5pt}
			Task & Method & Acc. at $100\%$ & Acc. at $80\%$ & Acc. at $60\%$ & AURC \\
			\hline
			A & QNN-E1 & 93.84 & 97.55 & 98.37 & 0.0137 \\
			A & QNN-E2 & 92.75 & 98.08 & 99.05 & 0.0145 \\
			A & CNOT All-to-All & 86.35 & 92.42 & 95.03 & 0.0406 \\
			A & Conjunctive fusion & \textbf{96.53} & \textbf{99.77} & \textbf{99.97} & \textbf{0.0028} \\
			B & QNN-E1 & 92.13 & 96.41 & 98.09 & 0.0203 \\
			B & QNN-E2 & 89.17 & 93.03 & 94.65 & 0.0569 \\
			B & CNOT All-to-All & 86.01 & 92.23 & 96.32 & 0.0378 \\
			B & Conjunctive fusion & \textbf{93.90} & \textbf{98.40} & \textbf{99.19} & \textbf{0.0100} \\
			\Xhline{1.5pt}
	\end{tabular}}
\end{table}

Selective classification makes the Task B gain more visible. At $80\%$ coverage, conjunctive fusion exceeds QNN-E1 by $2.22$ percentage points on Task A and $1.99$ points on Task B. Its gains over the CNOT-only model are $7.35$ and $6.17$ points. The fused MFQS also supplies uncertainty diagnostics. At $\sigma=0.28$, incorrect predictions have higher BetP entropy and nonspecificity on both tasks. Nonspecificity rises from $0.1022$ to $0.1263$ on Task A and from $0.1014$ to $0.1183$ on Task B. These quantities support rejection analysis in addition to the final BetP decision.

The training costs also differ. The proposed method trains two five-qubit evidence QNNs independently and appends a fixed conjunctive circuit. The CNOT-only model jointly optimizes both sources through a coupled ten-qubit objective. In statevector simulation, joint optimization uses a $2^{10}$-dimensional state, compared with separate $2^5$-dimensional states for independent training, and therefore increases memory and gradient-evaluation costs. The proposed fusion stage uses no linear-system solver or trainable fusion parameters. Actual runtime depends on the simulator or hardware backend; the experiment does not measure a runtime speedup.

The gate counts agree with Table~\ref{credal_level_com_t}. For two sources ($k=2$), conjunctive fusion of prepared general MFQSs uses $n$ Toffoli-type gates. A direct classical commonality update processes $2^n$ coordinates. The five-element experiment therefore uses a fixed five-gate intersection module and does not materialize a 32-coordinate evidence vector during fusion. BetP statistics can be estimated from measurement samples; reconstructing every output mass retains the readout cost discussed after Table~\ref{credal_level_com_t}.

The results show that source complementarity controls the benefit of fusion. In Task A, both sources remain accurate and make different errors under gate perturbations. Conjunctive fusion reinforces compatible focal-set evidence and consistently exceeds the single-source models. In Task B, the weaker frequency-domain source limits the gain at full coverage and can introduce less reliable masses. The method is therefore most effective when each source contributes useful evidence.

The selective results add a second finding. Conjunctive fusion gives the lowest AURC on both tasks and concentrates more errors in the rejected low-confidence subset. This reliability ordering comes from mass-level fusion followed by BetP, without a learned fusion classifier. The CNOT-only model also has a fixed fusion block, but it requires joint optimization of a coupled ten-qubit objective and provides no explicit focal-set relation. Conjunctive fusion supports independent source training and retains the interpretation $A\cap B$.

\section{Conclusion}
\label{con}

This paper establishes an element-qubit encoding for Dempster-Shafer structures. Focal sets become addressable computational-basis states, and measurement probabilities recover their belief masses. The resulting circuits implement belief readout, pignistic-level quantities, credal-level fusion and $\alpha$-junctions, marginalization, vacuous extension, and ballooning extension. For prepared MFQS inputs, CCR and the entire $\alpha$-junction use $O(n)$ element-wise gate modules. Their direct classical coordinate constructions update $2^n$ or $2^{2n}$ focal-set-indexed quantities. This comparison applies to the reasoning subroutine and excludes arbitrary state preparation and full tomography. In the multi-domain MNIST experiment, conjunctive MFQS fusion provides competitive clean accuracy, improved robustness, and better selective rejection than the CNOT-only All-to-All model under simulated gate-control noise.

The belief function framework gives quantum evidence a direct and interpretable organization. Operations on individual frame elements propagate to the evidence level without repeated materialization of the complete power-set vector. Focal-set intersection and BetP explain each fusion decision, and the experiments show stronger resistance to gate-control perturbations in the studied setting. Future work will extend these circuits to evidential machine learning and evaluate their preparation, compilation, and readout costs on quantum hardware.


\section*{Acknowledgment}
This work is partially supported by the National Natural Science Foundation of China (Grant No. 62373078).

{
\footnotesize
\bibliographystyle{elsarticle-num} 
\bibliography{mybibfile}
}




\end{document}